\documentclass[a4paper,preprint]{elsarticle}
\usepackage[nodots,nocompress]{numcompress}
\usepackage{graphicx,subfigure}
\usepackage{amsmath}
\usepackage{amssymb}
\usepackage{hyperref}
\usepackage{lscape}
\usepackage{times}
\usepackage{rotating}
\usepackage{setspace}
\biboptions{super}
\usepackage{color}
\usepackage[ruled,titlenotnumbered]{algorithm2e}

\begin{document}
\begin{frontmatter}

\title{Accuracy of MAP segmentation with hidden Potts and Markov mesh prior models via Path Constrained Viterbi Training, Iterated Conditional Modes and Graph Cut based algorithms.}

\author[label2]{Ana Georgina Flesia}
\author[label1]{Josef Baumgartner}
\author[label3]{Javier Gimenez}
\author[label4]{Jorge Martinez}

\address[label2]{Conicet at UTN- Universidad Tecnol\'ogica Nacional regional C\'ordoba\\ and Famaf- Facultad de Matem\'atica Astronom\'\i a y F\'\i sica,  Universidad Nacional de C\'ordoba.\\ Medina Allende s/n. Ciudad Universitaria, 5000. C\'ordoba, Argentina.}
\address[label1]{Facultad de Ingenier\'\i a, FCEFyN, Universidad Nacional de C\'ordoba, Argentina.\\
 Vélez Sarsfield 1611. Ciudad Universitaria, 5000. C\'ordoba. }
\address[label3]{Famaf - Facultad de Matem\'atica Astronom\'\i a y F\'\i sica,  Universidad Nacional de C\'ordoba.\\ Medina Allende s/n. Ciudad Universitaria, 5000. C\'ordoba, Argentina.}
\address[label4]{Departamento de Matem\'atica, Universidad Nacional del Sur,  Avenida Alem 1253. 2� Piso
Bah\'{\i}a Blanca, B8000CPB, Argentina.}

\begin{abstract}
Pixelwise image segmentation using Markovian prior models depends on several hypothesis that determine the number of parameters and general complexity of the estimation and prediction algorithms. The Markovian neighborhood hypothesis, order and isotropy, are the most conspicuous properties to set.

In this paper, we study statistical classification accuracy  of two different Markov field environments for pixelwise image segmentation, considering the labels of the image as hidden states and solving the estimation of such labels as a solution of the MAP equation
\begin{equation}
s^*=arg\displaystyle\max_{s}p(s|I,\theta),
\nonumber
\end{equation}
where $I$ is the observed image and $\theta$ the model. The emission distribution is assumed the same in all models, and the difference lays in the Markovian prior hypothesis made over the labeling random field. The a priori labeling knowledge will be modeled with a) a second order anisotropic Markov Mesh and b) a classical isotropic Potts model. Under such models, we will consider three different segmentation procedures, 2D Path Constrained Viterbi training for the Hidden Markov Mesh, a Graph Cut based segmentation for the first order isotropic Potts model, and ICM (Iterated Conditional Modes) for the second order isotropic Potts model.

We provide a unified view of all three methods, and investigate goodness of fit for classification, studying the influence of parameter estimation, computational gain, and extent of automation in the statistical measures Overall Accuracy, Relative Improvement and Kappa coefficient, allowing robust and accurate statistical analysis on synthetic and real-life experimental data coming from the field of Dental Diagnostic Radiography. All algorithms, using the learned parameters, generate good segmentations with little interaction when the images have a clear multimodal histogram. Suboptimal learning proves to be frail in the case of non-distinctive modes, which limits the complexity of usable models, and hence the achievable error rate as well.

All Matlab code written is provided in a toolbox available for download from our website, following the Reproducible Research Paradigm.

\end{abstract}

\begin{keyword}
Hidden Markov Models
\sep Hidden Potts Models
\sep EM for Gaussian mixtures
\sep Image segmentation
\sep Relative improvement\sep Confidence intervals for Kappa coefficient of agreement

\end{keyword}

\end{frontmatter}

\section{Introduction}

Bayesian approaches to image segmentation, classification and even compression have been devised since the early work on Causal Markov Mesh (CMM) of  Abend, Harley and Kanal \cite{Abend} in the sixties, and Hidden Markov random field (HMRF) models pioneered by Besag \cite{besag1974} in the seventies. Basically they consists of embedding the problem into a probabilistic framework by modeling each pixel as a random variable, with a given likelihood, embedding the knowledge of the hidden labels on a prior distribution. The key issue in using Markovian  models for segmentation applications is parameter estimation, which is usually a computationally expensive task. In practice, often a trade off is accepted between accuracy of estimation and running time of the estimation algorithm, see Chen et. al (2012) \cite{Chen2012}for an interesting review of the field.

A common approach consists of alternatively restoring the unknown segmentation based on a maximum a posteriori (MAP) rule and then estimating the model parameters using the observations and the restored data. For Random Markov Fields, there are available iterative approximated solutions  based on the use of the Gibbs distribution as a prior. This is the case, for instance, in the popular ICM algorithm of Besaj \cite{besag1986}, which makes use of pseudo-likelihood approximation. Complete Monte Carlo Markov Chain methods for Potts model and segmentation estimation were implemented on Pereyra et al. (2013) \cite{Pereyra2013}. In Gimenez et al. (2013) \cite{gimenez2013}, a new pseudo-likelihood estimator of the smoothness parameter of a second order Potts model was proposed and included on a fast version of Besaj's  Iterated Conditioned Modes (ICM) segmentation algorithm. Comparisons with other classical smoothness estimators, like the one introduced by Frery et al (2009) \cite{Frery2009}, were also provided. Levada et al (2010) \cite{Levada2010} also discussed MAP-MRF approximations combining suboptimal solutions based on several different order isotropic  Potts priors.

Direct competitors of ICM are the methods discussed by Celeux et. al (2003)\cite{celeux2003}  based on mean field like approximations to the expectation-maximization (EM) algorithm for Potts models with first order neighborhoods, which consider conditional probabilities rather than restorations of the hidden data. The EM algorithm is widely used in the context of incomplete data and in particular to estimate independent mixture models due to its simplicity. 

Binary segmentation problems, as background-object problems, have been addressed successfully with graph theoretical methods. Greig et al.(1989)\cite{greig1989} showed that graph cuts can be used for binary image restoration as solutions of the MAP-MRF problem. Boykov et al.(2001)\cite{Boykov2001} proposed a framework that used s/t graph cuts to get a globally optimal object extraction method for N-dimensional images. Blake et al.(2004) \cite{Blake2004} used a mixture of the Markov Gibss random field (MMGRR) to approximate the regional properties of segments and the spatial interaction between segments. The standard s/t graph cut algorithm can find the exact optimal solution for a certain class of energy functionals ; however, in many cases the number of labels for assigning to graph nodes is more than two,and the minimization of energy functions becomes NP-hard. For approximate optimization, Boykov et al.(2004)\cite{Boykov2004} developed the $\alpha$-expansion-move and $\alpha\beta$ swap move
algorithms to deal with multi-labeling problems for more general energy functionals.  An up to date survey of graph theoretical approaches to image segmentation is given in Peng et al (2013)\cite{Peng2013}.

Li et al.(2000)  \cite{li2000} introduced the first analytic solution to a true anisotropic 2-D hidden Markov model (HMM). They studied a strictly-causal, nearest-neighbor, 2-D HMM, and show that exact decoding is not possible. For block-based causal modeling, they assumed that both training and testing images shared the same underlying Markovian model up to transition probabilities.  The estimated the interblock transition probabilities with Path-Constrained 2-D Viterbi algorithms (PCVA) in training step.  Pyun et al (2007)\cite{Pyun2007} introduced a new method for blockwise image segmentation  combining the mode approximation method  as in Besaj \cite{besag1986}, with the simplest noncausal Markovian model, the bond-percolation model (BP) and its multistate extension (MBP). They did not assume homogeneity in spatial coherence among images, and estimated the hidden BP parameters in testing step using an stochastic EM procedure (mode field approximation) with computation time comparable to that using causal HMMs \cite{li2000} for blockwise segmentation of the same aerial images. Ma et al. (2010) \cite{ma2010} proposed a pseudo noncausal HMM by splitting the noncausal model into multiple causal HMMs, solved in a distributed computing framework for pixelwise segmentations. Other decoding approximations, also made on testing step, are discussed in Perronin et al {2003} \cite{Perronin2003},  Sargin et al. (2008) \cite{Sargin2008} and references therein.

As we said before, the key issue in using Markovian prior models for segmentation applications is the complexity of parameter estimation. Methods considering isotropic neighborhoods (classical Potts and bond-percolation models) can smooth out important quasi linear structures in images, while methods that consider non-isotropic neighborhoods, like causal 2dHMM, and non-isotropic Potts,  involve more parameters to estimate.
\begin{figure}[h!]
\centering
		\includegraphics[scale=.5]{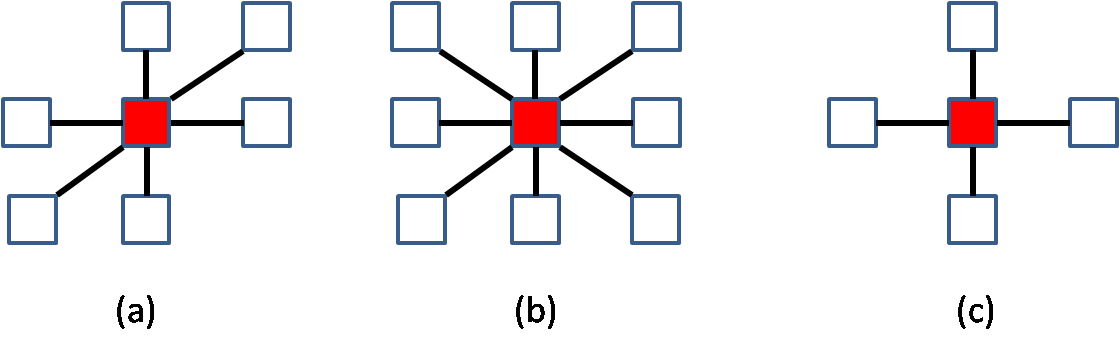}
\caption{Neighborhood systems: (a) 2D order Causal Markov Mesh , (b) second order Potts model, (c) first order Potts model. }
\label{vecinos}
\end{figure}

In this paper, we explore two specific Markov prior hypothesis for image segmentation, in the form of different neighborhood systems and probability relationships:
\begin{enumerate}
\item a diagonal six-pixel neighborhood for the anisotropic MRF
\item a four-pixel (first order) and eight-pixel (second order) neighborhood for the isotropic MRF
\end{enumerate}
which are depicted in Figure \ref{vecinos}.

The Markovian hypothesis assumed for each labeling Random Markov Field is different, in order to develop different estimation strategies. We keep the same Multivariate Gaussian model for the emission distributions. For the first neighborhood system we ensure assumptions of a second order Causal Markov Mesh model, which introduced an extra assumption in the neighborhood probabilities related to the notion of ``pixel's past''. In the second neighborhood system, we introduce a Gibbs distribution, in the form of the first and second order isotropic Potts model with smoothness parameter $\beta$.  Under such models, we will consider three different segmentation procedures,  2D Path Constrained Viterbi testing (PCVT) for the Hidden Markov Mesh, a simple proposal of Graph Cut (GC) based segmentation for the first order isotropic Potts model, and  Iterated Conditional Modes (ICM) for the second order isotropic Potts model. For all methods, we propose a unified framework consisting in three stages. The first stage is the image modeling, and initial labeling. The second stage is general parameter estimation and the third stage is computing the approximated MAP solution.

Previous work on causal 2dHMM for blockwise and pixelwise segmentation assume that both training and testing images share the same underlying Markovian model up to transition probabilities, \cite{li2000, Won2002, ma2010}. They estimate those probabilities in a training step and simply use them to test new images. However, this homogeneity assumption is arguable in many images. In our paper, we assume that the spatial coherence can vary from image to image, thus we estimate parameters and transitions probabilities in testing step. We also discuss specific choices for constraining paths possibilities in Viterbi decoding, that produces computation time comparable to that using ICM and GC.

In Sections \ref{MAP-rule}, \ref{gibbs-prior} and \ref{causal-prior} we discuss in detail the equations of the Markov Mesh prior model and Potts prior model, and the approximation choices made in our implementations of PCVT, GC and ICM. The goal of this paper is to give a unified view of these approximations, theoretically and algorithmically,  and compare their classification rates with synthetic and real data with available ground truth.  In Section \ref{experim} we introduce the design of our simulated experiments and the statistics used to evaluate classification. The implementation issues and prospects are given in Section \ref{summary}. All Matlab code written is provided in a toolbox available for download from our website, following the Reproducible Research Paradigm.

\section{MAP-MRF rules}
\label{MAP-rule}
Many  effective computational tools for practical problems in image processing and computer vision have been devised through Markov Random
Fields (MRF) modeling. One of these practical problems is to label an image domain pixelwise with given $L$ discrete labels $\mathcal{L}=\{\ell_1,...,\ell_L\}$, with the help of a priori modeling hypothesis like the Potts model. It corresponds to a special case of such Markov Random Field (MRF) over the image graph, which computes the graph partition with the minimal total perimeter of the one-label (constant) regions. It does not favor any particular order of the labels, in comparison to another similar model introduced by Ishikawa \ref{Ishikawa}, which partitions the image graph with multiple linearly ordered labels (overlapped to each other). The MRF-based energy function proposed by Potts model is the sum of the unary potential defined on each graph node and the pairwise potential given on each graph edge.

\subsection{Markov Random Fields}

Markov random fields provide convenient prior for modeling spatial interactions between pixels. Markov Random Fields were first introduced into vision by Geman and Geman \cite{geman1984}.

Let $\mathcal{P}$ be a set of pixels in the image $I$ of size $n=z\times w$,  $\mathcal{L}=\left\{\ell_{1},\ell_{2},\ldots,\ell_{L}\right\}$ a set of $L$ labels, and  for each site $(i,j)\in\mathcal{P}$ is defined as a set $\partial_{ij}\subset\mathcal{P}$ called the \emph{neighborhood} of $(i,j)$.
The neighborhood system is a collection of sets $\partial=\left\{\partial_{ij}:(i,j)\in\mathcal{P}\right\}$ which satisfies:
(a) $(i,j) \notin \partial_{ij}$,(b) $(i',j')\in \partial_{ij}$ $\Rightarrow$ $(i,j)\in \partial_{i'j'}$,
(c) $\mathcal{P}=\bigcup_{(i,j)\in\mathcal{P}} \partial_{ij}$.

The labeling problem is to assign a label from the label set $\mathcal{L}$ to each site in the set of sites $\mathcal{P}$. Thus a labeling is a mapping from $\mathcal{P}$ to $\mathcal{L}$. We will denote a labeling by $s=\left\{s_{ij}\right\}$. The set of all possible labeling $\mathcal{L}^{n}$ is denoted by $\mathcal{S}$.

In this paper we will consider the final segmentation $s=\left\{s_{ij}\right\}$ as  realizations of a Markov random field. This means that for each possible realization (called  configuration) $s\in\mathcal{S}$, it holds that
\begin{itemize}
\item $p\left(s\right) > 0 $,
\item $p\left(s_{ij}|s_{\mathcal{P}-\left\{(i,j)\right\}}\right) = p\left(s_{ij}|s_{\partial_{ij}}\right).$
\end{itemize}

where $\mathcal{P}-\left\{(i,j)\right\}$ denotes set difference, and $s_{\partial_{ij}}$ denotes the labels of the sites in $\partial_{ij}$.

\subsection{Maximum a Posteriori Estimation}

In general, the labeling field is not directly observable in the experiment. We have to estimate its realized configuration $s$ based on an observation $I$, which is related to $s$ by means of the likelihood function $p\left(I|s,\theta\right)$, where $\theta$ represents the set of all model's parameters. The most popular way to estimate an MRF is to maximize a posteriori (MAP) estimation. The MAP-MRF framework was popularized in vision by Geman and Geman \cite{geman1984}.

MAP estimation consists of maximizing the posterior probability $p\left(s|I,\theta\right)$. From the point of view of Bayes estimation, the MAP estimate minimizes the risk under the zero-one cost function. Using Bayes rule, the MAP estimate is
\begin{equation}
s^{*}=arg\max_{s\in \mathcal{S}}p\left(s|I,\theta\right)=arg\max_{s\in \mathcal{S}}p\left(I|s,\theta\right)p\left(s|\theta\right)
\label{s*}
\end{equation}
\subsection{Gaussian observation field}

For the sake of completeness, we consider multispectral images where each pixel is a vector from $\mathbb{R}^q$.  We assume that the observed pixel intensities are Multivariate Gaussian Mixtures, and the emission probabilities given the state $\ell\in\mathcal{L}$ are Multivariate Gaussian:
\begin{equation}
p(x \vert \ell)=\displaystyle\frac{1}{(2\pi)^{q/2}|\Sigma_\ell|^{1/2}}\exp\left\{-\displaystyle\frac12(x-\mu_\ell)^T\Sigma_\ell^{-1}(x-\mu_\ell)\right\}
\label{gaussianas}
\end{equation}
 with mean $\mu_\ell$ and covariance matrix $\Sigma_\ell$.
In this work, we compute the initial classification without considering any contextual prior, following Algorithm \ref{ML}.

\begin{algorithm}
\caption{Maximum Likelihood Classification}
\label{ML}
\begin{itemize}
\item[1)] Supervised Maximum Likelihood (ML) segmentation of $I$: To assign the label given by
$$s_{ij}=arg\max_{\ell\in\mathcal{L}}p(I_{ij}\vert \ell)$$
to pixel $(i,j)$, with parameters chosen from the modes of a histogram, or selected pixel values.

\item[2)] Unsupervised Maximum Likelihood (EM-ML) segmentation of $I$: To assign the label given by
$$s_{ij}=arg\max_{\ell\in\mathcal{L}}p(I_{ij}\vert \ell)$$
to pixel $(i,j)$, with parameters given by the Expectation Maximization algorithm for Gaussian Mixtures, with random initializations.
\end{itemize}
\end{algorithm}
The labeling obtained by ML will be the benchmark classification.

\section{Gibbs prior}
\label{gibbs-prior}
MRFs are generalizations of Markov processes that can be specified either by the joint distribution or by the local conditional distributions. However, local conditional distributions are subject to nontrivial consistency constraints, so the first approach is most commonly used. In this paper, we will consider two types of Markov random Fields as prior constraints for the labeling field, causal Markov Meshes, and Gibbs random fields.

Before defining Gibbs random fields  (GRF) we need to define a \emph{clique}. A set of sites is called a clique if each member of the set is a neighbor of all the other members. A Gibbs random field can be specified by the joint Gibbs distribution:
\[
p\left(s\right) = Z^{-1}\exp\left(-\sum_{C\in\mathcal{C}}V_{C}\left(s\right)\right),
\]
where $\mathcal{C}$ is the set of all cliques, $Z$ is the normalizing constant, and $\left\{V_{C}:C\in\mathcal{C}\right\}$ are real functions, called the clique potential functions. In this model, the conditional distribution of state label $s_{ij}\in\mathcal{L}$ corresponding to pixel $(i,j)\in\mathcal{P}$ given the evidence in the image is
\begin{equation}
p(s_{ij}|s_{i'j'}:(i',j')\in \partial_{ij})=p(s_{ij}|s_{i'j'}:(i',j')\neq(i,j))\propto \exp\left(-\sum_{C\in\mathcal{C}: (i,j)\in C}V_{C}\left(s\right)\right).
\label{porb}
\end{equation}
This Markovian assumption guarantees the existence of the joint distribution of the process.

\subsection{Potts model}

In this model,  the potential functions $V_C$ of (\ref{porb}) are defined as follows:
\begin{equation}
V_C(s)=\left\{\begin{array}{lll}
-\beta & \ & \mbox{if }s_{ij} = s_{i'j'},C=\{(i,j),(i',j')\}\in\mathcal{C},\\
0 & & \mbox{in other case.}
\end{array}
\right.
\label{potentials}
\end{equation}
where $\mathcal{C}$ is the clique set corresponding to the neighborhood's system $\partial$. Thus, the distribution on the neighborhood in the Potts model becomes
\[p(s_{ij}|s_{i'j'}:(i',j')\in \partial_{ij})\propto\exp\{\beta U_{ij}(s_{ij})\}\]
where $U_{ij}(s_{ij}):=\#\{(i',j')\in \partial_{ij}:s_{i'j'}=s_{ij}\}$, and $\beta$ is the smoothness parameter, sometimes called inverse temperature. Thus, the joint likelihood of the Markov random field is
\[p(s)\propto\exp\{\beta U_s\},\]
where $U_s=\#\{C\in\mathcal {C}:C=\{(i,j),(i',j')\},s_{ij}=s_{i'j'}\}$.

The observed process, which is supposed to be emitted by the hidden Markov Field, is considered Multivariate Gaussian as in  (\ref{gaussianas}) with mean $\mu_l$ and covariance matrix  $\Sigma_l$ which depends on the classes. Thus, given the observed pixel intensities $I$, the a posteriori distribution of the map of classes is
\begin{equation}
p(s|I,\theta)\propto\exp\{\beta U_s+\sum_{ij}p(I_{ij}|s_{ij})\}.
\label{energia}
\end{equation}
This distribution corresponds to a new Potts model in which the external field in a given pixel $(i,j)$ is $p(I_{ij}|s_{ij})$.

The optimum segmentation $s^*$ is defined  as a MAP solution (\ref{s*}), with $\theta=(\beta,\mu_l,\Sigma_l)$ which is a hard problem to solve in general.

There are many approximated solutions provided in the literature. In this paper, we will work with Iterated Conditional Modes and Graph Cut, algorithms that will be detailed in the following subsections.

A classical proposal for $\beta$ estimation consists in solving the maximum pseudolikelihood equation
\begin{equation}
\sum_{(i,j)\in \mathcal{P}}U_{ij}(s_{ij}) - \sum_{(i,j)\in \mathcal{P}}  \frac{\sum_{\ell\in L}U_{ij}(\ell)\exp\{\widehat{\beta} U_{ij}(\ell)\}} {\sum_{\ell\in L}\exp\{\widehat{\beta} U_{ij}(\ell)\}}=0.
\label{pmv}
\end{equation}
which involves an observed class map $s$, where the initial $s$ is given by maximum likelihood. Using this map, we calculate $U_{ij}(\ell)$ with $\partial_{ij}$ equal to the first or second order neighborhood, as depicted in Figure \ref{vecinos}. This assumption reduces equation (\ref{pmv}) to a smaller nonlinear equation in $\beta$, with coefficients counting the number of patches with certain configurations; see \cite{Levada2009} for details. This equation is solved in this paper using Brent's algorithm \cite{brent1973}.

\subsection{Iterated conditional Modes}

 Iterated Conditional Modes (ICM)  is an iterative algorithm that rapidly converges to the local maximum of the  function $P(s|I,\theta)$  closest to the initial segmentation provided by the user. Usually, the initial segmentation is provided by Maximum Likelihood. In each iteration  ICM modifies the label of each pixel  for the label that is most probable, given the neighborhood configuration.

Given observations $I$, ICM finds a suboptimal solution of (\ref{s*}), with the algorithm \ref{ICM}.
\begin{algorithm}
\caption{Iterated Conditional Modes (ICM)}
\label{ICM}
\begin{itemize}
\item[1)] Maximum Likelihood segmentation of $I$ (ML or EM-ML, see Algorithm \ref{ML}).
\item[2)] Parameter estimation by pseudo-maximum likelihood with Brent's algorithm for the smoothness parameter of the second order isotropic Potts model.
\item[3)] Choose a pixel's  visit scheme for the image.
\item[4)] For each pixel $(i,j)$, change the label given in the previous iteration for the  label $\ell\in\mathcal{L}$ that maximizes
\begin{equation}
g(\ell)=\ln p(I_{ij}|\ell,\hat{\mu}_{\ell},\hat{\Sigma}_\ell)+\hat{\beta}U_{ij}(\ell)
\label{maxICM}
\end{equation}
\item[5)] Iterate until convergence.
\end{itemize}
\end{algorithm}
The first term of (\ref{maxICM}) is equivalent to the ones used by the  ML classifier. The second term is the contextual component scaled by the parameter $\beta$.
If $\beta>0$, ICM smooths out the initial segmentation,if  $\beta<0$, ICM reduces clusters coherence. When $\beta=0$ the rule is reduced to the maximum likelihood, but when  $\beta\rightarrow\infty$ the effect is reversed, the data does not have any importance in the final segmentation.

\subsection{Graph Cut based approximation to MAP-MRF rules}

Energy based segmentation methods attempt to minimize energy functions, which can be obtained from several sources.
The general form of an energy function (notation taken from work by Boykov, Veksler and Zahib \cite{Boykov2001}) is as follows
\begin{equation}
\label{ecu1}
E\left(s\right) = \sum_{(i,j)\in\mathcal{P}} D_{ij}\left(s_{ij}\right)+\sum_{\left\{(i,j),(i',j')\right\}\in\mathcal{C}} V_{\left\{(i,j),(i',j')\right\}}\left(s_{ij},s_{i'j'}\right)
\end{equation}
where $D_{ij}$ is called the data penalty function and indicates the compatibility of the labeling with the given data. We can transform the Potts model equation  (\ref{energia}) into an energy based formulae
  \[
 p(s|I,\theta)\propto\exp\{-E(s)\}
 \]
 considering the first expression $D_{ij}\left(s_{ij}\right)$  as
\begin{equation}
D_{ij}\left(\ell\right)=-\ln p(I_{ij}|\ell,\hat{\mu}_\ell,\hat{\Sigma}_\ell)
\end{equation}
 and the second term, called the interaction potential or the smoothness function, as the density given by the first order Potts model.
 The smoothness term incorporates the notion of a piecewise smooth world and penalizes assignments that label neighboring nodes differently. The potential functions  $V_C$ are the ones given in (\ref{potentials}).

\begin{figure}[htb]
\centering
		\includegraphics[scale=0.1]{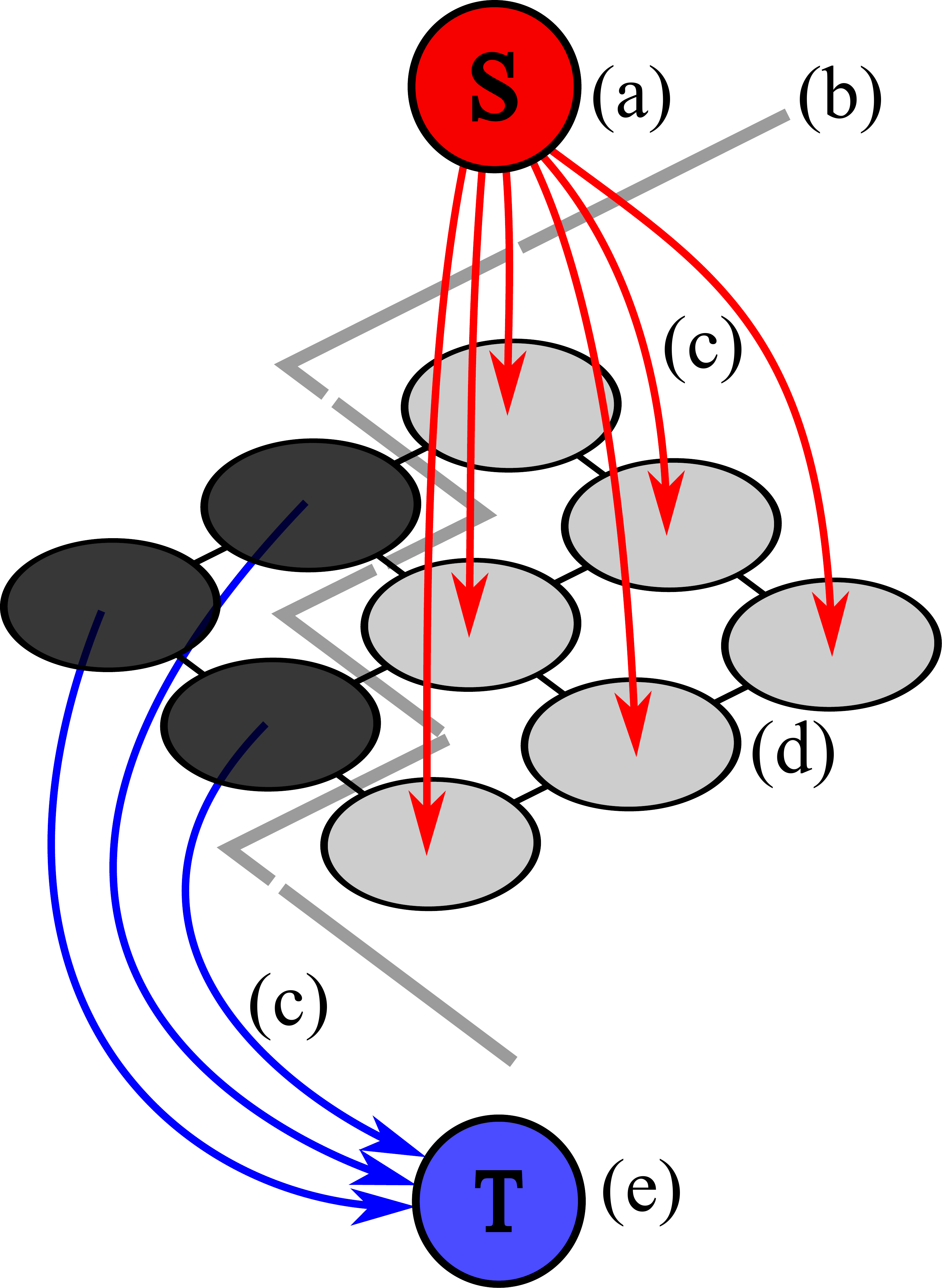}
		\caption{Example of flow network used in 2D image segmentation. (a) Source/Object (b) Cut  (c) T-links (d) N-links (e) Sink/background.}
\label{example_gc}
\end{figure}

General energy functions (\ref{ecu1}) can be minimized by a graph cut algorithm  \cite{Boykov2004},  \cite{Kolmogorov2004}. The image can be considered as a weighted graph $\mathcal{G}= \left\langle \mathcal{V},\mathcal{E}\right\rangle$, where the vertices's $\mathcal{V}$ are the pixels, and the edges $\mathcal{E}$ are the links between neighboring pixels. For a binary graph cut problem, two additional nodes known as source and sink terminals are added to the graph. The terminals correspond to the labels being assigned to the nodes, i.e., pixels of the image. A cut is a set of edges that separates the source and sink terminals such that no subsets of the edges themselves separate the two terminals. The sum of the weights of the edges in the cut is the capacity of the cut. In the implementation we used, the weights are the potential functions $V_C$, in other graph cut implementations, weights are defined as submodular functions. The goal is to find the minimum cut, i.e., the cut for which the sum of the edge weights in the cut is minimum.  Figure \ref{example_gc} is a representative diagram showing the process of partitioning the input image.
In the multiple label case, the multiway cut should leave each pixel connected to one label. This ensures that every multi-way cut which separates all terminals, must correspond to a valid labeling ie, a configuration of the associated MRF.  To solve the problem we used the $\alpha-$expansion move algorithm \cite{Boykov2001}. This algorithm minimizes an energy function with non binary variables by repeatedly minimizing an energy function with binary variables using Max-flow/Min-cut method. It starts with an arbitrary labeling and performs iterative optimization cycles until the process converges. Each cycle consists of iterating over the set of labels, running the  $\alpha-$expansion move once for every label $\alpha$. This involves finding a new labeling $s^\prime$ obtained by increasing the number of $\alpha$ labels, which is better than the current labeling $s$, ie. $E\left(s^\prime\right)\leq E\left(s\right)$. The algorithm will converge when in a particular cycle, no better $s^\prime$ can be found.

Given observations $I$, Graph Cut finds a suboptimal solution, with the following algorithm:
\begin{algorithm}
\caption{Graph-Cut (GC)}
\label{GC}
\begin{itemize}
\item[1)] Maximum Likelihood segmentation of $I$ (ML or EM-ML, see Algorithm \ref{ML}).
\item[2)] Parameter estimation by pseudo-maximum likelihood with Brent's algorithm, for the smoothness parameter of the first order isotropic Potts model.
\item[3)] Minimize (\ref{ecu1}) using an $\alpha-$expansion move algorithm, using the initial segmentation ML and the $\hat\beta$ estimated in step 2).
\end{itemize}
\end{algorithm}

\section{Causal prior}
\label{causal-prior}
For pixel $(i,j)$ we define the following causal relationship that represents the ``past'', $ (i^{'},j^{'}) \prec (i,j) $ if $ i^{'}<i $ or $ i^{'}=i $ and $ j{'}<j $, and the set $\{s_{i,j-1} , s_{i-1,j}\}$ the neighborhood of $(i,j)$ in the past.  We assume a Causal 2D order Markov Mesh model stating that, for state $s_{ij}$:
\begin{equation}
p(s_{ij} \vert s_{i'j'}: (i',j') \prec (i,j)) = p( s_{ij} \vert s_{i,j-1} , s_{i-1,j}).
\label{relacion_vecinos_HMM}
\end{equation}
The two pixels $(i,j-1)$ and $(i-1,j)$ can be understood as the \textquotedblleft past\textquotedblright \ of pixel $(i,j)$. We consider $P(s_{ij} \vert s_{i,j-1} , s_{i-1,j})$ to be independent of the current pixel so we can gather the transition probabilities in a matrix $A$ where
$$a_{\ell_1,\ell_2,\ell_3}=p(s_{ij}=\ell_1\vert s_{i,j-1}=\ell_2 , s_{i-1,j}=\ell_3) \qquad \forall \ell_1,\ell_2,\ell_3\in\mathcal{L}.$$

The transition matrix $A$ has one more dimension than the transition matrix of a 1d Markov Process due to two ``past states'' on the left and above the actual pixel. This yields at a new order of the image. Instead of lining up the pixels as we would have done in the one-dimensional case we are now moving from the top-left pixel to the bottom-right pixel. Thus, the initial probabilities for the 2dMM depend only on the first state $s_{0,0}$ and we can write
\[\pi_\ell=p(s_{0,0}=\ell) \qquad \forall \ell\in\mathcal{L}.\]
The word ``hidden'' that is usually added to the whole model (Gaussian observed process plus Markov Mesh labeling random field) comes from the fact that this Markov Mesh can not be observed, so it is considered hidden. It can be proved that this causal relationship implies a general Markov Field hypothesis with the diagonal neighborhood stated in the introduction, that is, the probability of a label given the whole labeling specification depends only on the values in the pixels depicted in Figure \ref{vecinos} (a).

\begin{figure}[h!]
\centering
		\includegraphics[scale=.6]{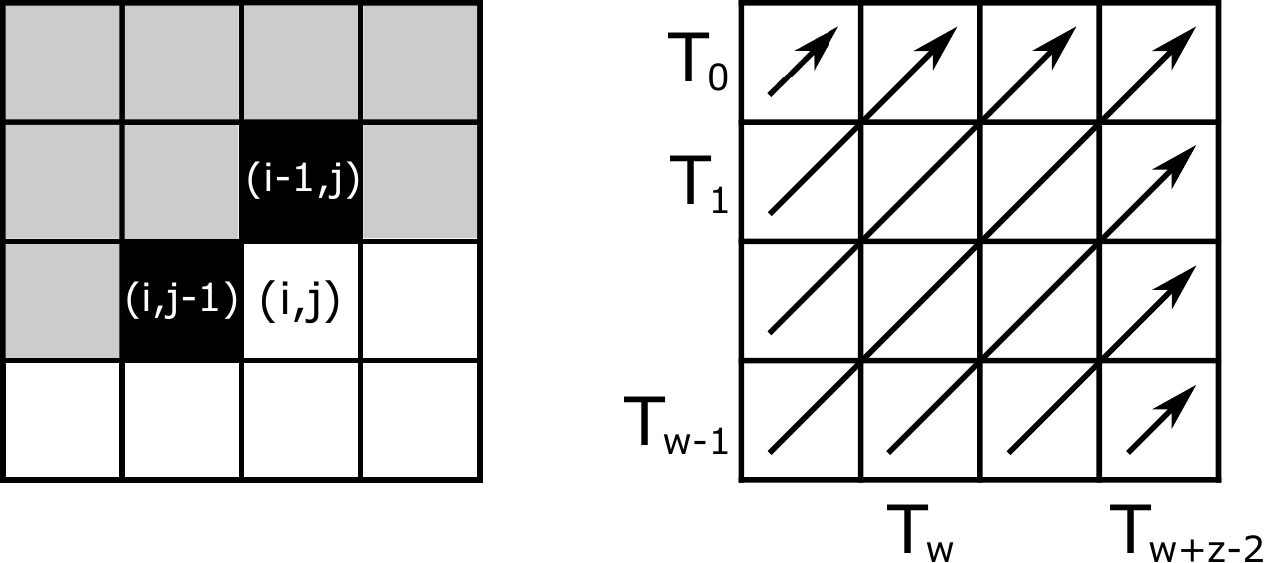} 
		\caption{Transitions among states in a 2nd order Markov Mesh and Pixels on diagonals $T_0,\ldots,T_{w+z-2}$}
		\label{pasado}
\end{figure}
If we enumerate each diagonal in the image, $T_0,\ldots,T_{z+w-2}$, as one step in time, starting with the top-left pixel, see Figure \ref{markovt},
\[T_0=(s_{0,0}); \ \ T_1=(s_{1,0},s_{0,1}); \ \ T_2=(s_{2,0},s_{1,1},s_{0,2}); \ \ \ldots \ \ ; \ \ T_{z+w-2}=(s_{z-1,w-1});\]
 the Markov Mesh assumption (along with the particular definition of the past) implies that
\begin{eqnarray}\nonumber
p(s)&=&p(T_0)p(T_1|T_0)\ldots p(T_{z+w-2}|T_{z+w-3},\ldots,T_0) \\ \nonumber
&=& p(T_0)p(T_1|T_0)\ldots p(T_{z+w-2}|T_{z+w-3}).
\label{markovt}
\end{eqnarray}
This means that each diagonal operates as an \textquotedblleft isolating\textquotedblright \ element between neighboring diagonals, which suggest an extension of the 1d Viterbi algorithm to compute the most probable sequence of states given initial values. This is, to find the optimal combination of states $s^*$ that solves (\ref{s*}) given the whole 2D hidden Markov model, the labeling field and the observed Gaussian intensity process.

Let  $T_0=(s_{0,0});\ \ T_1=(s_{1,0},s_{0,1});  \ \ \ldots$ be  a path through the image where every diagonal marks one step. Each diagonal consists of up to $\min(w,z)$ states: $T_0\in\mathcal{L}, \ T_1\in\mathcal{L}^2, \ T_2\in\mathcal{L}^3, \ \ldots , \ T_{z+w-2}\in\mathcal{L}$. This makes  a total of $L^{\min(w,z)}$ possible state combinations only considering the main diagonal. Therefore, the exact decoding of our problem is an NP-hard problem. To produce an approximated solution we will work constraining the set of possible state combinations.

\subsection{Path-Constrained Viterbi Training}

The Viterbi training algorithm is an iterative algorithm that estimates all the parameters of a HMM, and finds the sequence of states that best explains the data, given the estimated parameters. The procedure starts with the setting of initial parameters, which can be done using prior information, educated guess or a non-contextual estimation.  Using this initial step the algorithm follows the next steps until convergence
\begin{algorithm}
\label{PCVT}
\caption{Path-Constrained Viterbi Training (PCVT)}
\begin{itemize}
\item Initialize segmentation $s^{(0)}$: Maximum Likelihood segmentation of $I$ (ML or EM-ML, see Algorithm \ref{ML}).
\item Parameter estimation:  Given  sequence $s^{(n-1)}$,  estimation of $a_{\ell_1,\ell_2,\ell_3}^{(\mathfrak{n})}$, $\mu_\ell^{(\mathfrak{n})}$ and $\Sigma_\ell^{(\mathfrak{n})}$
\item Decoding: Choosing the best $N$ paths and Viterbi decoding using these paths.
\end{itemize}
\end{algorithm}

\subsubsection{First step Viterbi Training: Parameter estimation}

Let's suppose we have the initial sequence $s^{(0)}$ obtained from Algorithm \ref{ML}, or the sequence  $s^{(n-1)}$ obtained from Viterbi in the previous step. Our empirical estimations of the transition probabilities and distributions parameters are

\begin{equation}
a_{\ell_1,\ell_2,\ell_3}^{(\mathfrak{n})}=\displaystyle\frac{\displaystyle\sum_{i=1}^{z-1}\displaystyle\sum_{j=1}^{w-1}\chi\left(s_{i-1,j}^{(\mathfrak{n}-1)}=\ell_1,s_{i,j-1}^{(\mathfrak{n}-1)}=\ell_2,s_{ij}^{(\mathfrak{n}-1)}=\ell_3\right)}
{\displaystyle\sum_{i=1}^{z-1}\displaystyle\sum_{j=1}^{w-1}\chi\left(s_{i-1,j}^{(\mathfrak{n}-1)}=\ell_1,s_{i,j-1}^{(\mathfrak{n}-1)}=\ell_2\right)}
\label{vitrain_a2}
\end{equation}
\begin{equation}
\mu_\ell^{(\mathfrak{n})}=\displaystyle\frac{\displaystyle\sum_{i=0}^{z-1}\displaystyle\sum_{j=0}^{w-1}\chi\left(s_{ij}^{(\mathfrak{n}-1)}=\ell\right)I_{ij}}{\displaystyle\sum_{i=0}^{z-1}\displaystyle\sum_{j=0}^{w-1}\chi\left(s_{ij}^{(\mathfrak{n}-1)}=\ell\right)}
\label{vitrain_mu2}
\end{equation}
\begin{equation}
\Sigma_\ell^{(\mathfrak{n})}=\displaystyle\frac{\displaystyle\sum_{i=0}^{z-1}\displaystyle\sum_{j=0}^{w-1}\chi\left(s_{ij}^{(\mathfrak{n}-1)}=\ell\right)(I_{ij}-\mu_\ell)(I_{ij}-\mu_\ell)^T}{\displaystyle\sum_{i=0}^{z-1}\displaystyle\sum_{j=0}^{w-1}\chi\left(s_{ij}^{(\mathfrak{n}-1)}=\ell\right)}
\label{vitrain_sigma2}
\end{equation}
where $\chi$ is the indicator function.

\subsubsection{Second step Viterbi training: Decoding}
\label{decoding}
There are several different approximations in the literature for iterative decoding. Sargin et al (2008)\cite{Sargin2008} proposed an algorithm that iteratively updates the posterior distribution on rows and columns, i.e. determining  horizontal and vertical 1d  forward-backward probabilities, combining them to approximate the values of $p(s_{ij}\vert s_{i,j-1}, s_{i-1,j})$ as product of horizontal and vertical probabilities. A more simplistic approach is to represent the dependency of the neighbors as the horizontal and vertical conditionals, a row and column wise constrained application of belief propagation. Such models deviate us from the original Makovian assumptions, so in this paper we will follow the so called Path Constrained Viterbi Training Algorithm, Li et al (2000)\cite{li2000}, Ma et al (2010)\cite{ma2010},  which restricts the possibilities of diagonal strings of states to propose a labeling, and updates all parameters in a pseudo-Expectation Maximization way using such labeling until convergence. We will describe now the equations involved in the process.
\begin{figure}[htb]
\centering
		\includegraphics[scale=0.4]{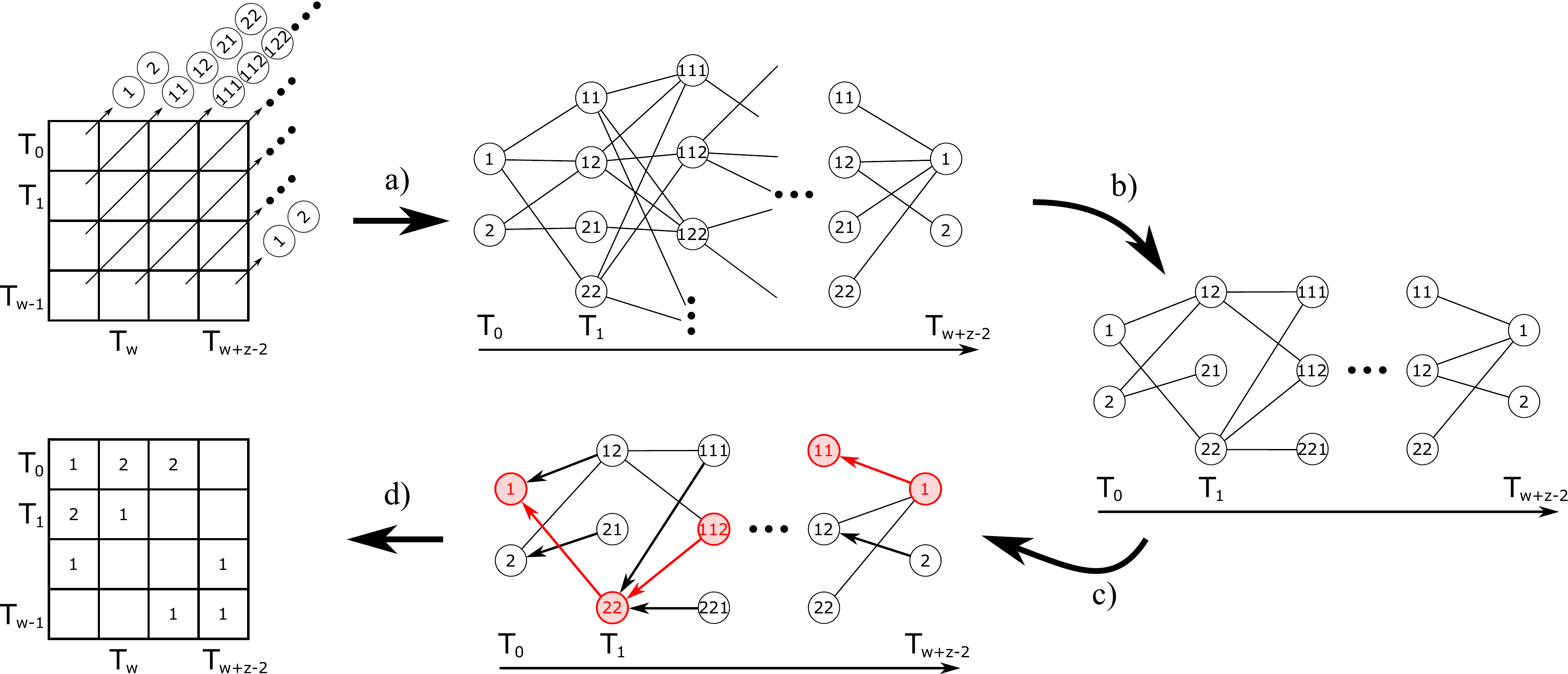}
		\caption{\textsl{Viterbi Decoding} for two possible states: a) Viterbi transformation. b) Path constrained to $N=3$. c) Tracking back the optimal path of state sequences using the values saved in $\varphi$. d) Hidden state map.}
		\label{Viterbi}
\end{figure}

\begin{itemize}

\item {\bf Choosing the best $N$ paths for decoding}

In this paper, we propose the following selection. We  firstly assume, as Li et al (2000) \cite{li2000},  that we can evaluate the likelihood of a given diagonal state sequence by simply multiplying the likelihoods of each pixel without considering statistical dependencies between pixels, i.e.
we compute
\begin{equation}
\hat p(s_{ij}=\ell|I_{ij},\hat\theta)\propto p(I_{ij}|s_{ij}=\ell,\hat\theta)\hat p(s_{ij}=\ell|\hat\theta)
\label{esta}
\end{equation}
where $p(I_{ij}|s_{ij},\hat\theta)$ is given by (\ref{gaussianas}) and
     \[\hat{p}(s_{ij}=\ell|\hat\theta)=\displaystyle\frac{\displaystyle\sum_{i=0}^{z-1}\displaystyle\sum_{j=0}^{w-1}\chi(s_{ij}=\ell)}{zw} \ \ \ \ \forall \ell\in\mathcal{L}.\]
Thus, the most likely state sequence $\textbf{s}_{d,1}$ is the one that has in each entry the most likely state for the pixel's observation on diagonal $d\in\{0,1,\ldots,z+w-2\}$.

In our particular implementation, we will obtain the next $N-1$ sequences considering only the sequences that result from changing only one state of $\textbf{s}_{d,1}$. Such chains are ordered using (\ref{esta}) and the $N-1$ with the largest likelihood are chosen.

In our Discussion section we will comment the incidence of the selection of this bag of $N$ sequences in the convergence of our implementation.

\item {\bf Viterbi decoding over the chosen $N$ paths.}
In Figure \ref{Viterbi} we show an schematic example of the decoding algorithm, which follows the idea of the original 1d Viterbi algorithm, using diagonals as pivotal points.
We call each diagonal state sequence $\mathbf{s}_{d,k}$ where $d$ is the index for the diagonal with $d=0,1,\ldots,z+w-2$ and $k=1,2,\ldots,N$ indicates the state sequence. Hence the initial state probabilities $\tilde{\pi}_k$ for pixel $(0,0)$ are
$$\tilde{\pi}_k=p(T_0=\mathbf{s}_{0,k}).$$
We denote $\delta_d(k)$ as the maximum probability for sequence $k$ on diagonal $d$. Given the parameters of the PCVT we can write
$$\delta_d(k)=\displaystyle\max_{k_{0},k_{1},\ldots,k_{d-1}}{p(\mathbf{s}_{0,k_0},\ldots,\mathbf{s}_{d-1,k_{d-1}},\mathbf{s}_{d,k},\mathbf{I}_0,\ldots,\mathbf{I}_d|\theta)}, $$
with $d=0,\ldots,z+w-2;  k=1,\ldots,N.$
Furthermore we collect the pixels on diagonal $d$ in a variable $\Delta(d)$ and define
$$b_{\mathbf{s}_{d,k}}(\mathbf{I}_d)=\displaystyle\prod_{(i,j)\in\Delta(d)}p(I_{ij}|\mathbf{s}_{d,k}(i,j))$$
where $\mathbf{I}_d=(I_{ij}:(i,j)\in\Delta(d))$ and $b_{\mathbf{s}_{d,k}}(\mathbf{I}_d)$ is the emission probability of sequence $k$ on diagonal $d$ under the assumption that each pixel is statistically independent from its neighbors. Finally, we can calculate the transition probability from sequence $k$ on diagonal $d$ to sequence $l$ on diagonal $d+1$:
$$\tilde{a}_{d,k,l}=p(T_{d+1}=\mathbf{s}_{d+1,l}|T_d=\mathbf{s}_{d,k},\theta)=\displaystyle\prod_{(i,j)\in\Delta(d+1)}a_{\mathbf{s}_{d,k}(i-1,j),\mathbf{s}_{d,k}(i,j-1),\mathbf{s}_{d+1,l}(i,j)}$$
$$d=0,\ldots,z+w-3; \ \ \ k,l=1,\ldots,N.$$
Now we are ready to initialize the \textsl{Viterbi decoding Algorithm} with the values
$$\delta_0(k)=p(T_0=\mathbf{s}_{0,k}), \quad b_{\mathbf{s}_{0,k}}(\mathbf{I}_0)=\tilde{\pi}_k b_{\mathbf{s}_{0,k}}(I_{0,0}) \qquad  \forall k=1,2,\ldots,N.$$\\
Then we start the recursion
$$\delta_{d+1}(l)=\biggl[\displaystyle\max_{1 \leq k \leq N}{\delta_d(k)\tilde{a}_{d,k,l}}\biggr]b_{\mathbf{s}_{d+1,l}}(\mathbf{I}_{d+1}) \qquad \forall d=0,1,\ldots,z+w-3 \quad \forall l=1,2,\ldots,N.$$
After each step, we save the index of the most probable sequence on diagonal $d$ that leads to sequence $l$ on diagonal $d+1$ in a variable called $\varphi$:
$$\varphi_{d+1}{(l)}=\arg\displaystyle\max_{1 \leq k \leq N}\{\delta_d(k)\tilde{a}_{d,k,l}\} \qquad \forall d=0,1,\ldots,z+w-3 \quad \forall l=1,2,\ldots,N$$
When the algorithm reaches the last diagonal, we use the values saved in $\varphi$ to track back the most probable path through the image starting with the bottom-right pixel
$$s_{z+w-2}^{*} = \arg\displaystyle\max_{1 \leq k \leq N}{\delta_{z+w-2}(k)}$$
$$s_{d}^{*}=\varphi_{d+1}{(s_{d+1}^{*})} \qquad  \forall d=z+w-3,z+w-4,\ldots,1$$
The final result $s^*$ contains the optimal path through the $N$ sequences at each diagonal. Note that this is equal to knowing the complete hidden state map for the whole image.
\end{itemize}

\section{Experimental Results: Image Segmentation}
\label{experim}
In this section we report some experiments on the three algorithms described in this article: Path Contrained Viterbi Training (PCVT, Algorithm \ref{PCVT}), Iterated Conditional Modes (ICM, Algorithm \ref{ICM}), and Graph Cut (GC, Algorithm \ref{GC}). For comparison purposes, we also provide the results when applying supervised (ML) or unsupervised (EM-ML) Maximum Likelihood Classification, with Algorithm \ref{ML}.
In the supervised case ML is initialized with the real means and variances, known from the simulation parameters, or taken from the GT in the case of  real images. In the unsupervised experiment, ML selects the EM initial parameters  with the modes of the histogram, and in the case of unimodal distributions, starting from random means. ML is the only algorithm among the tested ones that does not take into account spatial information. All the other algorithms were initialized with (ML) or (EM-ML) outputs, and run until convergence.

The PCVA code we made to carry out these experiments was designed from scratch, on a Matlab 2013a platform.  We used Matlab Statistical toolbox scripts for (EM-ML) and (ML). In the literature, initialization has also been made with non-parametric segmentation algorithms like k-means, while the means and variances for the Gaussian hypothesis on the observations were estimated over the labeled output. For the studied examples, we did not observe significant differences using k-means as initialization method.

\begin{figure}[h!]
\centering
  \includegraphics[scale = 0.3]{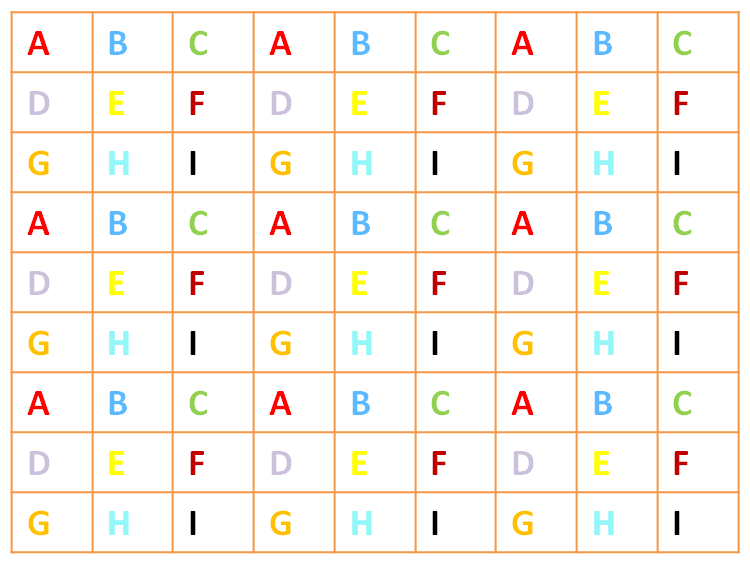}
  \caption{Algorithm for visiting sites adopted by ICM. All 9 cycles in one iteration are shown in different colors; the sites at each cycle are updated together. }
  \label{fig_visita}
\end{figure}

We implemented a version of ICM where parameter $\beta$ is estimated at each iteration by maximizing the current pseudo-likelihood with the Brent algorithm. The ICM visiting scheme consists in dividing the image support into $3\times3$ windows, and updating the labels of each cycle together as we can see in Figure \ref{fig_visita}. This visiting scheme minimizes convergence delays introduced by oscillations between sites.
For the Graph Cut algorithm,  the MAP solution is found as the minimum cut on a transportation network, using Boykov and Kolmogorov (2004) \cite{Boykov2004} available C code,  (MAXFLOW - software for computing mincut/maxflow in a graph, Version 3.01). This is also a benchmark code, widely used for segmentation of multimodal images with few classes \cite{Ali2007,Ali2008}.  The  $\beta$  parameter is estimated initially by maximizing the current pseudo-likelihood with the Brent algorithm, particularized to the order of the neighborhood system.

One of the goals of this paper is to quantify the gain in classification accuracy produced by imposing Markovian hypothesis to the labeling field. The performance is evaluated with the following statistical measures: Overall Accuracy, Kappa and Relative Improvement. The Kappa statistic is defined as
\begin{equation*}
\widehat{\kappa}=\frac{\sum_{\text{i}=1}^Lp_{\text{ii}}-\sum_{\text{i}=1}^Lp_{\text{i}+}p_{+\text{i}}}{1-\sum_{\text{i}=1}^Lp_{\text{i}+}p_{+\text{i}}}.
\end{equation*}
where  $p_{\text{ij}}$, are the proportion of pixels of the true class $l_{\text{j}}$ assigned to class $l_{\text{i}}$ in the segmented image,
$n$ is  the total of pixels in the image and $p_{\text{+j}}$ are the proportion of pixels of class $l_{\text{j}}$ and $p_{\text{i+}}$ are the proportion of pixels assigned to class $l_{\text{i}}$. $\widehat{\kappa}$ is an estimator of the degree of matching in a classification scheme. It helps to determine if the scheme output is better than random allocation. Fleiss, Cohen and Everitt \cite{testkappa} have shown an asymptotic variance for this estimator,
\begin{equation}\label{var_k_noind1}
\hat{\sigma}^2_{\hat{\kappa}}=\frac{\sum_{\text{k}=1}^Lp_{\text{kk}}[1-(p_{+\text{k}}+p_{\text{k}+})(1-\hat{\kappa})]^2+(1-\hat{\kappa})^2
\sum_{\text{j}\neq\text{k}}p_{\text{jk}}(p_{+\text{j}}+p_{\text{k}+})^2-[\hat{\kappa}-\sum_{i=1}^L p_{i+}p_{+i}(1-\hat{\kappa})]^2}{n(1-\sum_{i=1}^L p_{i+}p_{+i})^2},
\end{equation}
 that allows to define an asymptotic normal $100(1-\alpha)\%$ confidence interval for Kappa,  $\left(\hat{\kappa}-z_{\alpha/2}\hat{\sigma}_{\hat{\kappa}} \ , \ \hat{\kappa}+z_{\alpha/2}\hat{\sigma}_{\hat{\kappa}}\right)$.
The Overall accuracy index is defined by
\begin{equation*}
OA=\sum_{\text{i}=1}^Lp_{\text{ii}}.
\end{equation*}
It computes the proportion of coincidences between the reference and reconstructed images (well classified pixels). A classification scheme is considered good if its Overall Accuracy is above 85\%.
When appropriate, we also report the Relative Improvement index related to the benchmark  ML or EM-ML classification, defined as
\[
Relative\;Improvement=\frac{OA_{method}-OA_{ML}}{100-OA_{ML}}\times 100
\]

We investigated several examples. We first compared the performance of the algorithms in terms of the quality of the segmentations on simulated and real data with two classes, (Section{\ref{circulos}). Later on, we consider more complex scenarios studying scanned intra oral X-ray images with hand made Ground Truth (Section \ref{multiclass}).

\subsection{Two-class examples.}
\label{circulos}

Digitalized X-ray images have some level of noise introduced by the scanner, but their main characteristic is the smoothness of the joint gray level histogram. Classes that are quite distinguishable to the naked eye do not form a distinctive mode in the joint histogram, making segmentation difficult. Image subtraction, image enhancement and filtering are common image processing research areas when working with digitalized or digital X-ray imagery. Caution has been advised by dentists \cite{flesia-flesia} about the abuse of enhancement algorithms in digital X-ray devises, that often introduce artifacts in the images by defect and by excess, leading to possible misdiagnosis. The image data analyzed in this paper has been kindly provided by Odontologist J.G. Flesia, from the Image Processing section of the Dentistry department at University of C\'ordoba, who also helped us to judge the discrimination capabilities of our algorithms.

\subsubsection{Titanium screw}
Our first image is a scanned  bitewing X-ray image that shows a typical implant, consisting of a titanium screw (resembling a tooth root) and background bone tissue. Bitewing X-rays are used to detect decay between teeth and changes in bone density caused by gum disease. They are also useful in determining the proper fit of a crown (or cast restoration) and the marginal integrity of fillings.

The particularity of this image, Figure \ref{implant} (a), is its bimodal histogram Figure \ref{implant} (b). The segmentations of the observed image after applying the different algorithms are given in Figure \ref{implant} (c)-(f). This case is interesting in that it illustrates that all algorithms can obtain good unsupervised segmentations on real images when the image has a bimodal histogram, without any need of pre-filtering or enhancement.  PCVT slightly better delineates the right side of the implant that looks more fused to the bone.
\begin{figure}[htb]
\centering
     \subfigure[]{
  \includegraphics[scale = 0.5]{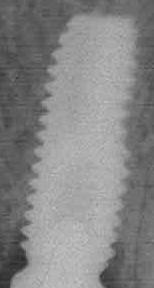}
 }\subfigure[]{
  \includegraphics[scale = 0.2]{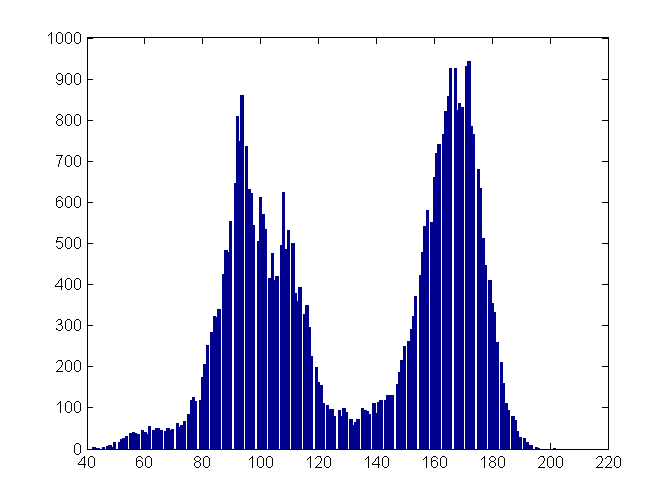}
 }\subfigure[]{
  \includegraphics[scale = 0.4]{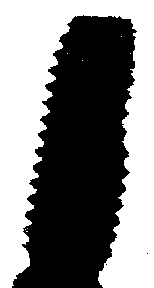}
 } \subfigure[]{
  \includegraphics[scale = 0.4]{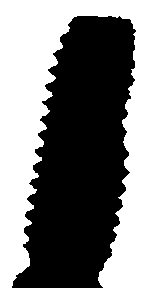}
 }\subfigure[]{
  \includegraphics[scale = 0.4]{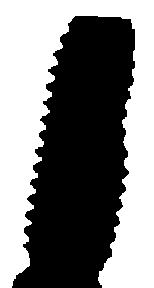}
 }\subfigure[]{
 \includegraphics[scale = 0.4]{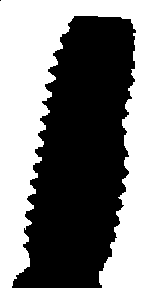}}
\caption{Inverse digitalized dental X-ray image which shows a titanium implant screw surrounded by bone. (a)original image, (b) histogram, (c) EM-ML segmentation, (d) ICM segmentation, (e) GC segmentation and (f) PCVT segmentation. }
 \label{implant}
\end{figure}

\subsubsection{Two-circles}

Here we discuss 40 experiments made with a synthetic $241\times241$ two color image to test the procedures ability to segment noisy images with unimodal and bimodal mixture distributions. Unimodal distributions are simulated by filling each class with Gaussian data with high variances; modes start to emerge when the variances are reduced.  Examples of the segmentations obtained are shown in Figure \ref{two-circles}. The corresponding parameters and performance statistics are given in Tables 3 and 4 as an indication of the algorithms ability to restore the truth in addition to visual assessment. These problems are difficult, so the algorithms using spatial information clearly outperform EM for independent mixture model in terms of restoration in both cases, supervised and unsupervised.  Moreover, the hidden Potts algorithms, ICM and GC, produce notably lower error rates than PCVT. Confidence intervals for Kappa values mostly revel that segmentations are statistically different, and that a good choice for the initial means in the supervised unimodal problems are essential to guarantee convergence in PCVT and GC. This case also illustrates that, when images generated from severely overlapping mixtures are considered, the PCVT performance decreases as the noise increases. This occurs to a lesser extent for ICM and GC,  since in this particular situation the use of MAP classifications turns out to be an advantage.

We also provide results for same problems when the noisy image is smoothed out by filtering, allowing the modes to emerge. In this case, the performance is boosted in all algorithms, surpassing 95\% accuracy and 0.9 Kappa value, Figure \ref{two-circles}, third row.

\begin{figure}[h]
\centering
	\[\begin{array}{cccccc}
\mbox{Hist} &\mbox{Original } &\mbox{EM-ML} & \mbox{EM-ML-ICM}&\mbox{EM-ML-GC}& \mbox{EM-ML-PCVT}\\
\includegraphics[width=2.8cm]{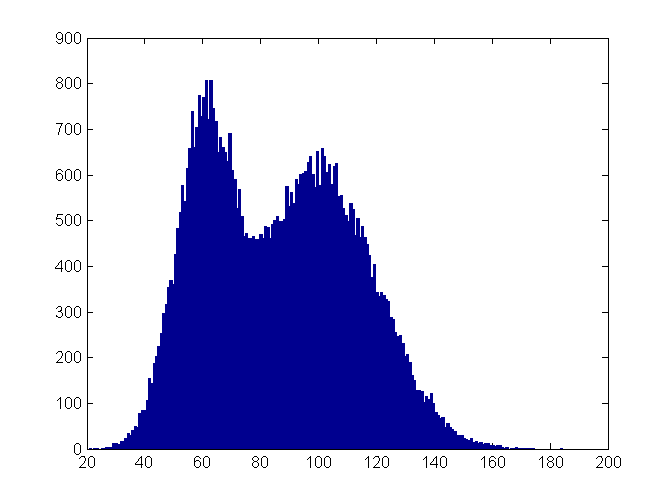} &\includegraphics[width=2cm]{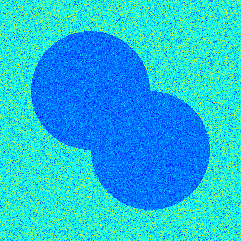} &\includegraphics[width=2cm]{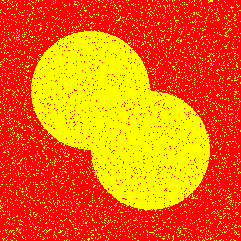}&\includegraphics[width=2cm]{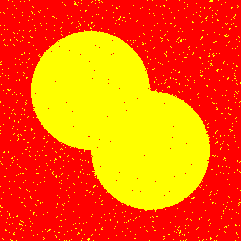}
&\includegraphics[width=2cm]{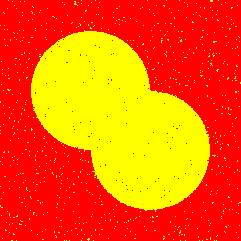}&\includegraphics[width=2cm]{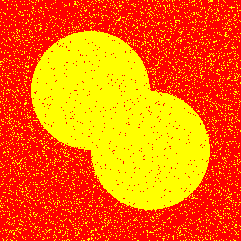}\\
\mbox{Hist} &\mbox{Original } &\mbox{ML} & \mbox{ML-ICM}&\mbox{ML-GC}& \mbox{ML-PCVT}\\
\includegraphics[width=2.5cm]{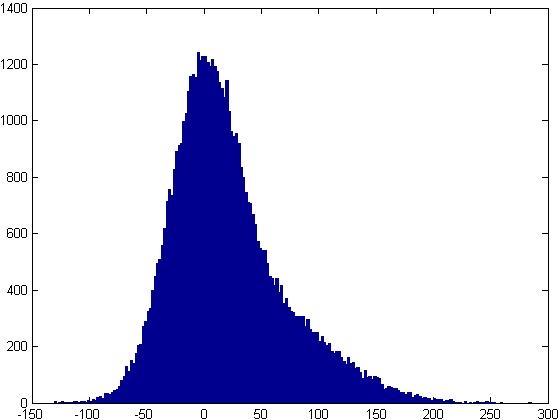} &\includegraphics[width=2cm]{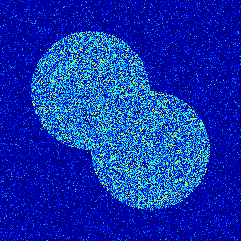} &\includegraphics[width=2cm]{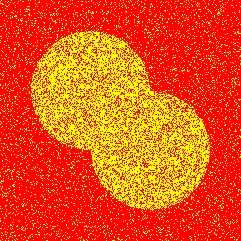}&\includegraphics[width=2cm]{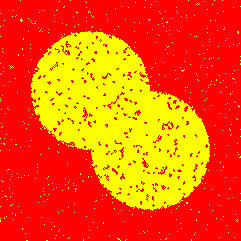}
&\includegraphics[width=2cm]{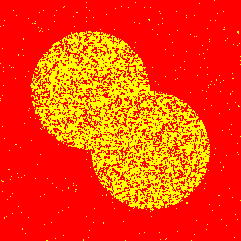}&\includegraphics[width=2cm]{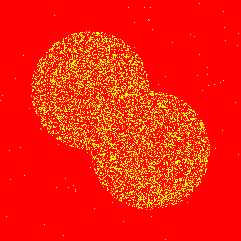}\\
\mbox{Hist} &\mbox{Filtered} &\mbox{EM-ML} & \mbox{EM-ML-ICM}&\mbox{EM-ML-GC}& \mbox{EM-ML-PCVT}\\
	\includegraphics[width=3cm]{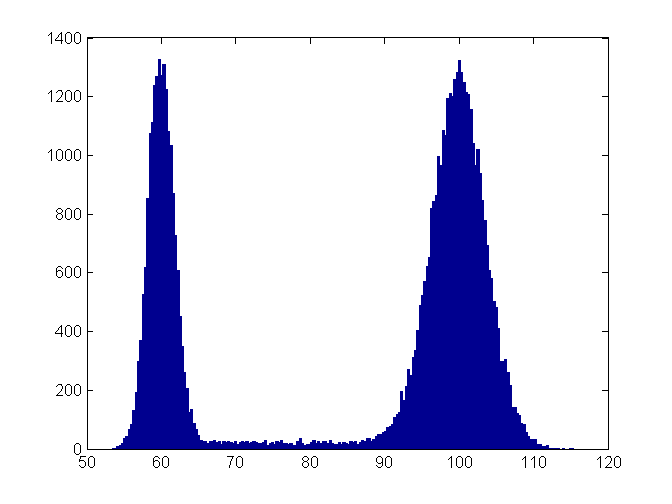} &\includegraphics[width=2cm]{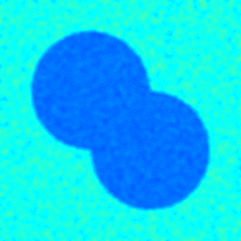} &\includegraphics[width=2cm]{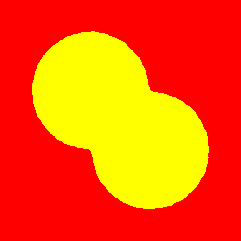}&\includegraphics[width=2cm]{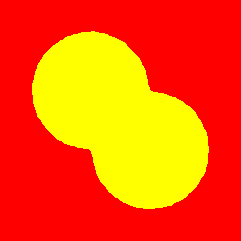}
&\includegraphics[width=2cm]{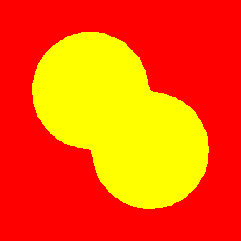}&\includegraphics[width=2cm]{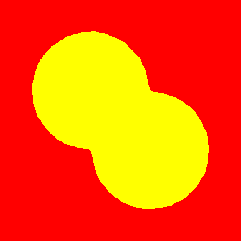}\\	 \end{array}	\]
	\caption{First row and second row: Noisy synthetic image, its histogram and segmentations. Third row, Unimodal synthetic image filtered, its histogram and unsupervised segmentations. }
	\label{two-circles}
\end{figure}

\subsubsection{Logo image}
Bidimensional Hidden Markov models are quite interesting models for the prior labeling field, with Gaussian mixture observations. Nevertheless, the complexity of the final equations calls for reductions and approximations with frail final estimations. In this section we want to discuss the influence of the selection of the best $N$ sequences for decoding in execution time, overall performance. Previous work with PCVT did not explicitly discuss these points, Ma et al (2010)\cite{ma2010} and Li et al (2000)\cite{li2000} give no particulars about the level of complexity added to the algorithm by the decision on how to select the bag of possible sequences, and the influence in the segmentation output of such decision. Besides, they reduced the number of computations estimating parameters and transition probabilities in the training stage, and using such information in testing new images.
 \begin{figure}[h!]
\subfigure[]{
 \includegraphics[scale = 0.3]{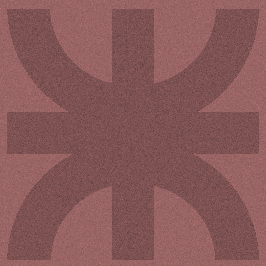} }
 \subfigure[]{
 \includegraphics[scale = 0.3]{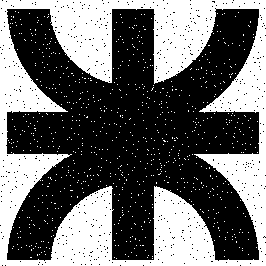} }
 \subfigure[]{
 \includegraphics[scale = 0.3]{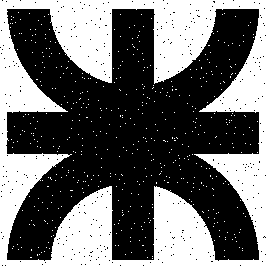} }
 \subfigure[]{
 \includegraphics[scale = 0.3]{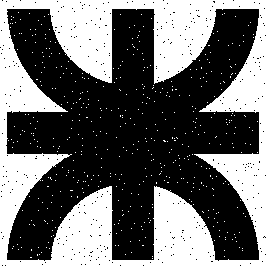} }
 \subfigure[]{
 \includegraphics[scale = 0.3]{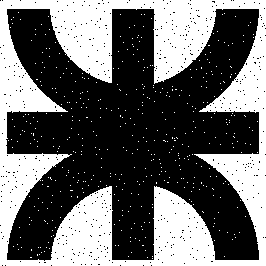} }
 \subfigure[]{
\includegraphics[scale = 0.35]{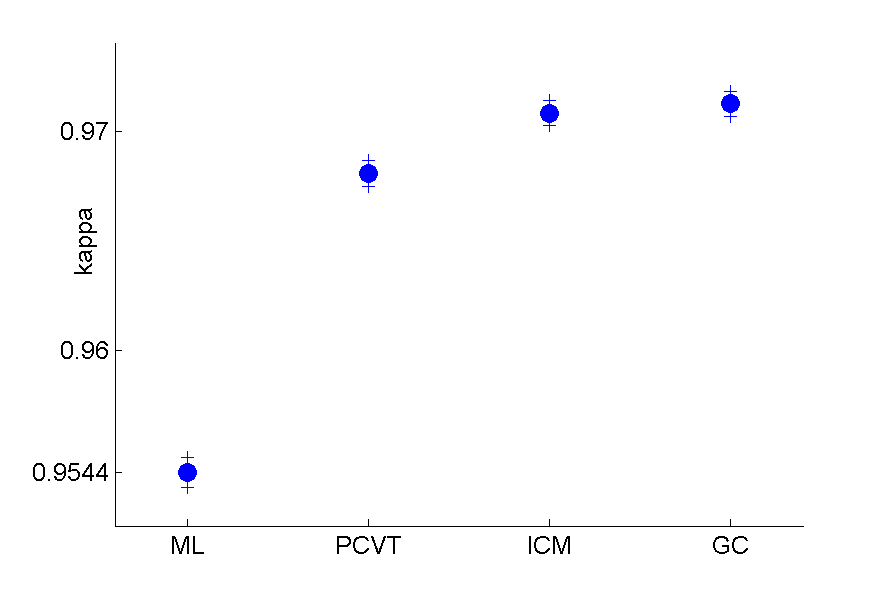}}
\subfigure[]{
 \includegraphics[scale = 0.35]{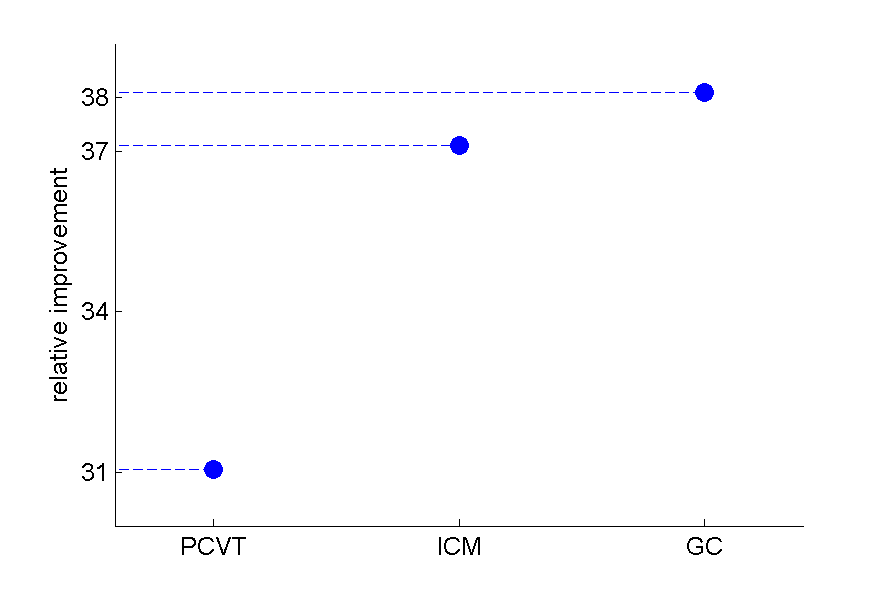} }
\subfigure[]{
 \includegraphics[scale = 0.35]{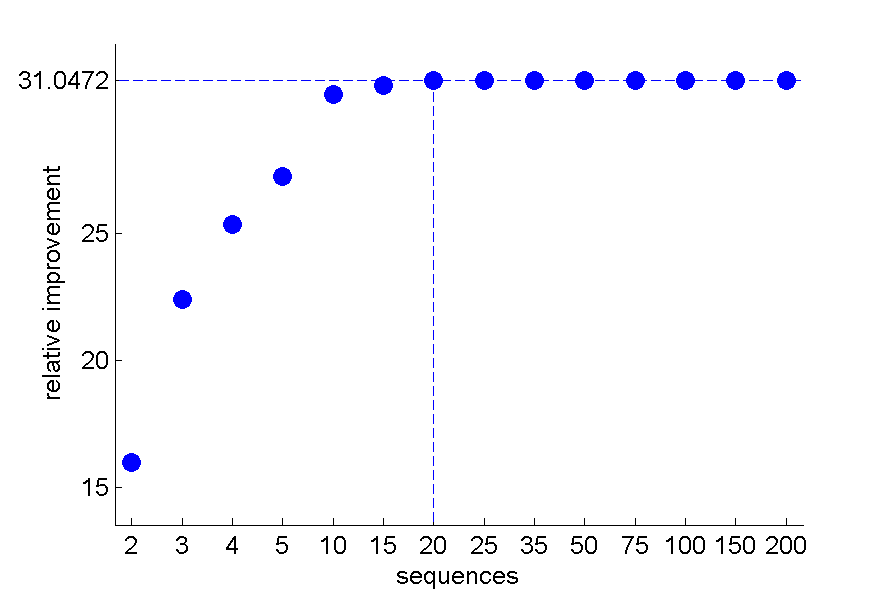} }
 \caption{(a) Noisy UTN logo, (b)EM-ML segmentation, (c)EM-ICM segmentation, (d)EM-GC segmentation, (e) EM-PCVT segmentation, (f) confidence intervals for kappa statistic, (g) Relative improvement of all methods related to Maximum Likelihood classification .(h) Relative improvement of PCVT related to ML vs number of sequences retained.}
   \label{logo}
  \end{figure}
 Our personal implementation allows the user to set the number of sequences $N$ involved in decoding, being 250 the preset value. We worked out a small  supervised study with the 2-color logo of the Technological University degraded with gaussian noise. We used Gaussian densities with class-dependent means being the true noise parameters $(\mu_1,\sigma1)=(60,15)$  and $(\mu_2,\sigma2)=(40,15)$. We made 16 experiments setting the number of path sequences allowed for decoding in a range from 1 to 250, as we can see in Table \ref{logo-imp}. We also computed time until convergence, number of iterations until convergence and relative improvement of classification accuracy related to ML segmentation.

This study shows that allowing the most probable 20 sequences has the same relative improvement as working with the most probable 250, and the time of execution goes from 0.8 minutes to 59 minutes on an Intel I7 processor, 6Gb memory HP laptop(see Table \ref{logo-imp}). We also show in this table that the number of iterations stabilizes when 50 or more sequences are allowed. This means that the algorithm uses the resources to search for a better minimum, but it was not found in the 250 sequences.

For the same image, the confidence intervals for Kappa also show that GC, PCVT and  ICM are significantly different from ML, see panel (c) of Figure \ref{logo}. The relative improvement of ICM and GC over ML were 37.1096 and 38.0894 respectively, see panel (d) of Figure \ref{logo}.

\begin{table}[h!]
\centering
\begin{tabular}{|c|c|c|c|c|}
  \hline
Number of& Number of&&Relative\\
Sequences& Iterations&Time&Improvement\\   \hline
1 &     2 &0.014729 & 0\\
2 &    10 &0.13618&15.9829\\
3 &    16 &0.23855 & 22.4127\\
4 &    23 &0.3619 &25.3521\\
5 &    22 &0.35359&27.2505\\
10&    25 &0.46362&30.4960\\
15&    28 &0.60956& 30.8634\\
20&    30 &0.80751&  31.0472\\
25&    30 &1.0695 &    31.0472\\
35&    30 &1.777&   31.0472\\
50&    36 &4.0193 &   31.0472\\
75&    36 &8.2936&   31.0472\\
100&   36 &13.606 &   31.0472\\
150&   36 &28.7398&   31.0472\\
200&   36 &46.1519  &   31.0472\\
250&   36 &59.2679 & 31.0472\\
  \hline
  \end{tabular}
 \caption{Relative improvement over ML, number of sequences, number of iterations, and time to convergence. Maximum number of iterations permitted, 100. }
 \label{logo-imp}
\end{table}
\subsection{Multiclass images. }
\label{multiclass}
\subsubsection{Multiclass high-quality bitewing X-ray image.}
\label{diente}

\begin{figure}[htb]
\centering
     \subfigure[]{
  \includegraphics[scale = 0.2]{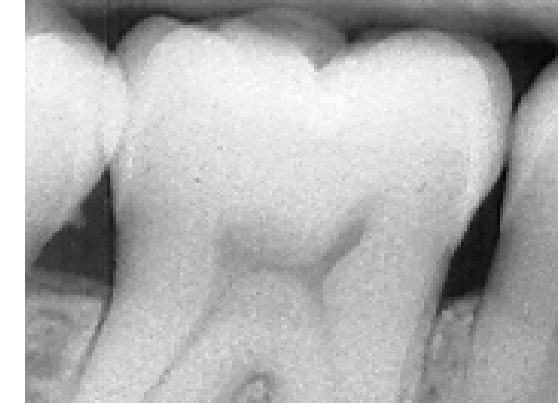}
 } \subfigure[]{
  \includegraphics[scale = 0.2]{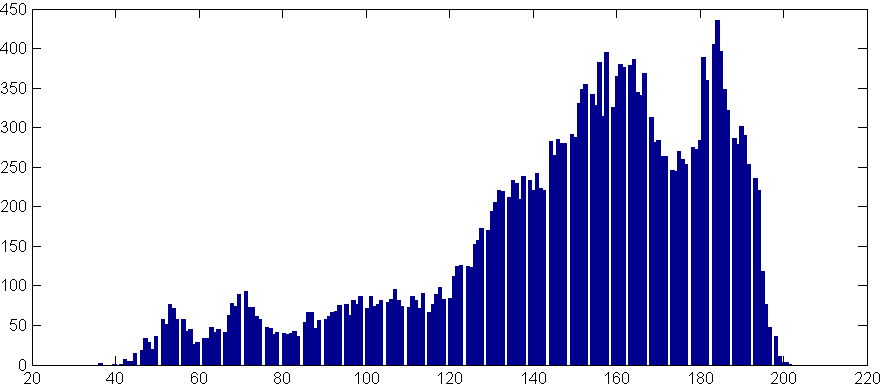}
  }

  \subfigure[]{
  \includegraphics[scale = 0.2]{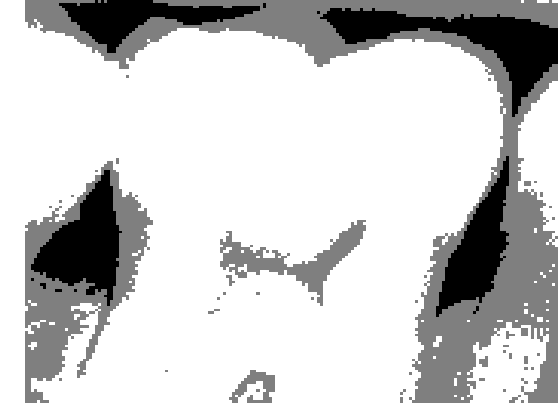}
 }
\subfigure[]{
\includegraphics[scale = 0.2]{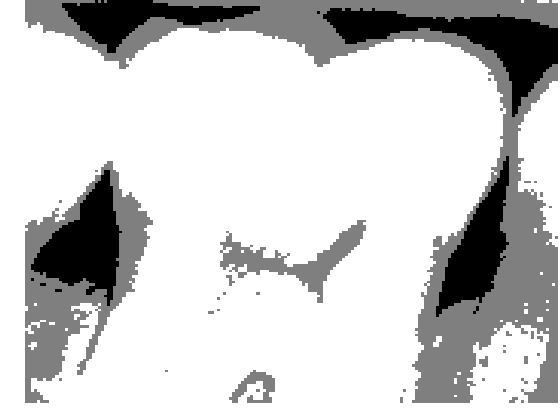}
}
 \subfigure[]{
 \includegraphics[scale = 0.2]{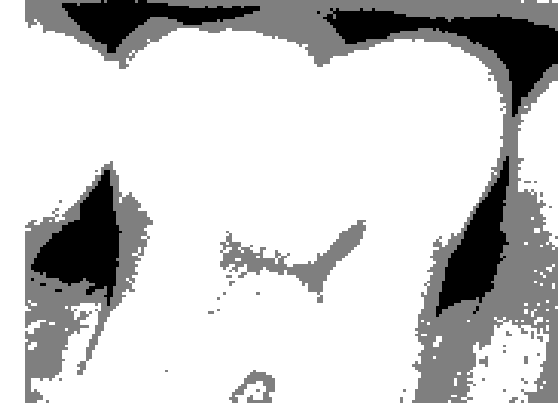}
 }\subfigure[]{
\includegraphics[scale = 0.2]{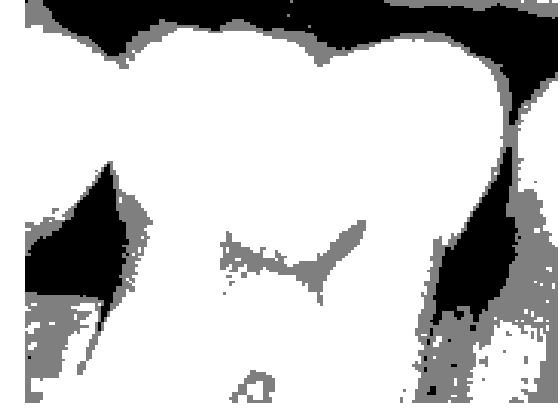}
 }

  \subfigure[]{
  \includegraphics[scale = 0.2]{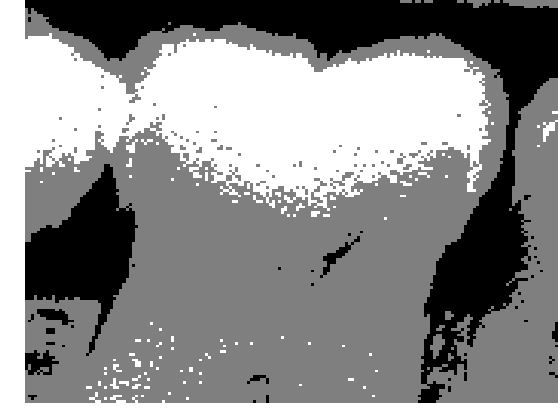}
 } \subfigure[]{
  \includegraphics[scale = 0.2]{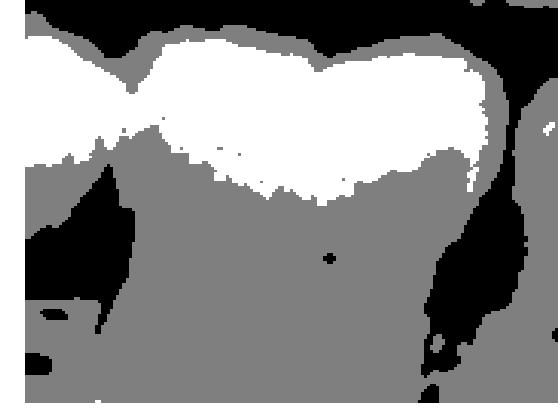}
 }\subfigure[]{
  \includegraphics[scale = 0.2]{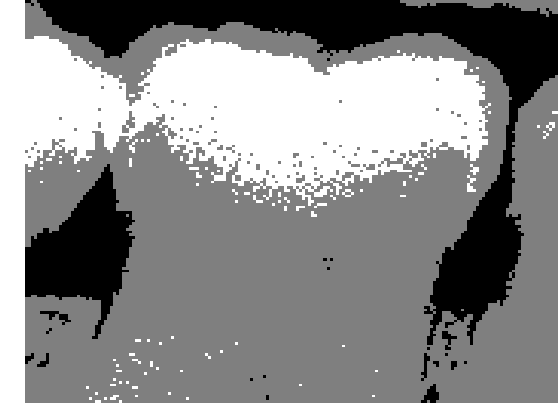}
 }\subfigure[]{
 \includegraphics[scale = 0.2]{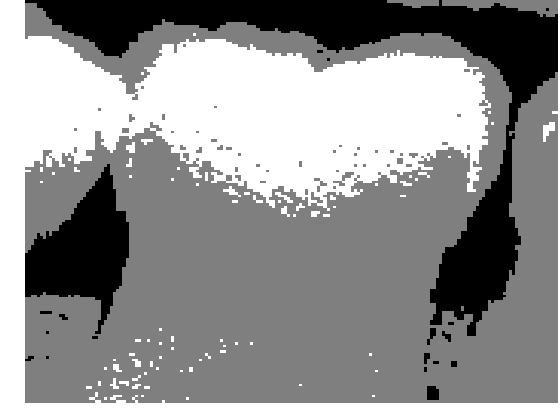}
   }
\caption{Segmentation of inverse digitalized dental X-ray image that has a central molar tooth and  partial views of its neighbor teeth, gums and background. First row, algorithms initialized with the modes of the histogram, fixed variance, second row, with automatic EM for Gaussian Mixtures. In both rows, from left to right, ML, ICM, GC and PCVT.}
 \label{molar}
\end{figure}
Our second  scanned  bitewing X-ray image shows a central molar tooth and  partial views of its neighbor teeth, gums and background, Figure \ref{molar} (a). We have set three classes in this image to account for the differences between tooth enamel and dentin. Enamel is the thin, hard material that covers the dentin, or main body of the teeth, and protects it from harsh temperatures. We initialized both algorithms with automatic EM and supervised classification. In Figure \ref{molar} we can observe from the segmentations that, in the supervised case, (c), (d), (e),  the teeth are correctly segmented, dentine is clearly differentiated from tooth nerve and enamel. 
Initializing with automatic EM, Figure \ref{molar} ((f), (g), (h), gives only a binary division, background from teeth and gum. In this example, PCVT makes a slightly better job than ICM, since the background is better separated from the teeth. The joint histogram of the image, Figure \ref{molar} (b), does  show at least three distinctive modes; in Figure \ref{incisivo}, the histogram shown in panel (a)corresponding to the X-ray image of an incisive,  is flat.  Its corresponding segmentations, shown in panels (c)-(f), made with four classes, do not distinguish well tooth enamel, dentin and background. In the expert's opinion, PCVT is the only method that separates correctly enamel from background and dentin.

\begin{figure}[h!]
\centering

 \subfigure[]{
\includegraphics[width=5cm,height=2.5cm]{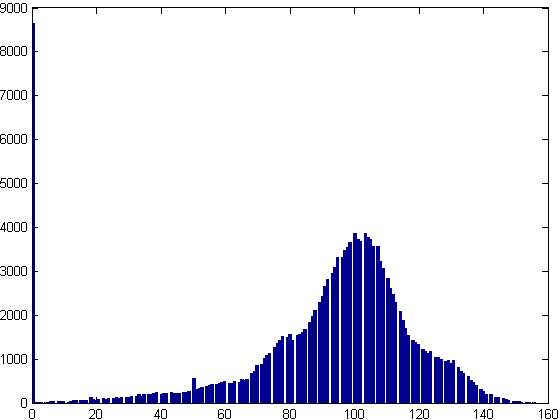}
 }
 \subfigure[]{
\includegraphics[scale = 0.2]{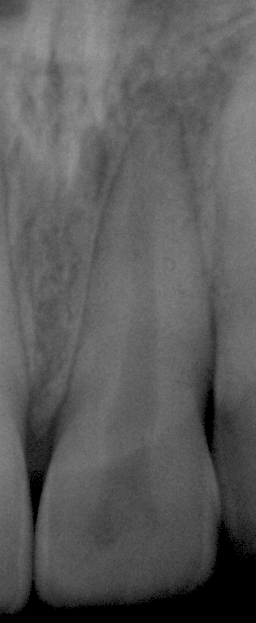}
 }
 \subfigure[]{
\includegraphics[scale = 0.2]{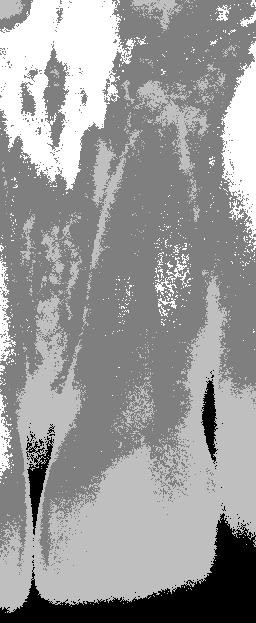}
 }\subfigure[]{
\includegraphics[scale = 0.2]{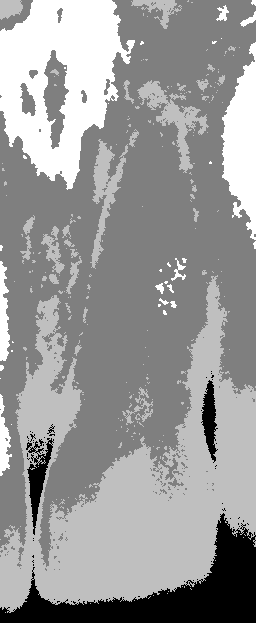}
 }
 \subfigure[]{
\includegraphics[scale = 0.2]{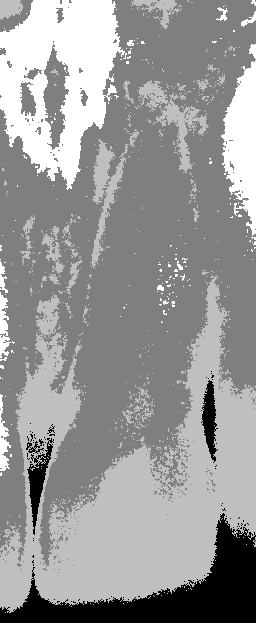}
 }\subfigure[]{
\includegraphics[scale = 0.2]{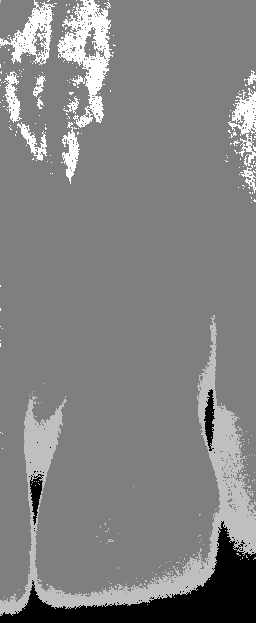}
 }
\caption{Segmentation of inverse digitalized dental X-ray images. Panels (a) incisive image (b) histogram of pixel intensities. From left to right, ML, ICM, GC and PCVT.  }
\label{incisivo}
\end{figure}
\subsubsection{Multiclass low quality X-ray image: Winstar rat jaw}
\label{rata}
In this section we work with imagery obtained from the Image Processing section of the Dentistry department at University of C\'ordoba. The rat jaw samples are part of a  response of bone tissue study, from pups with mothers stressed with continuous light conditions \cite{Fontanetti}. Histological studies were conducted with the samples to confirm bone loss. Since these studies destroy the samples, film based conventional dental X-ray images were carefully taken and digitalized with a flatbed scanner so as to document the study.

The image used here was segmented by Odontologist J.G. Flesia in 8 different classes. Pups were three days old, so the image's gray scale histogram only shows two main classes, background and jaw, after filtering.  In older rats, bone loss is documented comparing the clear region (soft parts) shown in experiment (c) of Figure \ref{seg_Luz} with the same region of the Control group. In few days old rats, the early bone inserted in the soft part is undeveloped and very difficult to segment. Under the advice of the experts, we considered the following experiments with this data:

 \begin{enumerate}

 \item[]Experiment (a): Two classes, complete jaw sample and  background.

 \item[]Experiment (b): Two classes,   teeth and jawbone from the rest of the image.

  \item[]Experiment (c): Three classes,  hard parts (teeth and jawbone),  soft parts (flesh and cartilage) and background.

   \item[]Experiment (d): Four classes, hard parts (teeth and jawbone), flesh, cartilage and background.

   \end{enumerate}

In Table \ref{taluz} we report Kappa and Overall accuracy for the four experiments. The confidence intervals give different conclusions for them. In experiment (b) and (c) Graph Cut has the highest Kappa value and it is statistically different from the others.  In experiment (a), all results are statistically different, and ICM has the highest Kappa value, followed by Graph Cut. In experiment (d), PCVT has the highest Kappa value, followed by Graph cut and ICM, being all statistically different.

PCVT does not have the best performance, but it reduces the background noise much better than the other methods, while maintaining ML accuracy. This is a striking feature detected also in the two-circles synthetic image. With PCVT, background was perfectly recovered, all errors were introduced in the object. The solution to this problem could be to consider subclasses inside main regions. Segmentation Problems (a) to (c) are contained in Problem (d), thus PCVT rates can be compared merging the confusion matrix of problem (d) as needed. Doing so we can conclude that PCVT works much better when the number of classes is closer to reality. GC, instead starts, having convergence problems, and its performance is better when two classes are involved. ICM is the only method that is not influenced by the number of classes considered.


\begin{table}[h!]
\centering
\begin{tabular}{|c|c|c|c|c|}
  \hline
  &\multicolumn{4}{|c|}{experiment (a)} \\
	\hline
	&ML &ICM &GC &PCVT \\
	\hline
	OA & 0,9633 & 0,9649 & 0,9631 & 0,9576  \\
	kappa & 0,9031$\pm$ 0.0004 &{\bf 0,9065$\pm$ 0.0004} & 0,9016$\pm$ 0.0004 & 0.8867$\pm$ 0.0004\\
	\hline
	&\multicolumn{4}{|c|}{experiment (b)} \\
	\hline
	&ML   &ICM	&GC  &PCVT	\\
	\hline
	OA    & 0,9699 & 0,9700 & 0,9710  & 0,9421 \\
	kappa & 0,8912$\pm$ 0.0005  & 0,8918$\pm$ 0.0005 & {\bf 0,8945$\pm$ 0.0006 }   & 0,8230$\pm$ 0.0004\\
	\hline
  &\multicolumn{4}{|c|}{experiment (c)} \\
	\hline
	&ML   &ICM	&GC &PCVT	 \\
	\hline
	OA    & 0,9231 & 0,9398 & 0,9442 & 0,9442 \\
	kappa & 0,8300$\pm$ 0.0005& 0,8596$\pm$ 0.0005 & {\bf 0,8672$\pm$ 0.0005} & 0,8606$\pm$ 0.0005   \\
	\hline
	&\multicolumn{4}{|c|}{experiment (d)} \\	
	\hline
	&ML 	  &ICM	&GC  &PCVT \\
	\hline
	OA& 0,8815 & 0,8983 & 0,9153  & 0,9246 \\
	kappa& 0,7573$\pm$ 0.0005 &0,7834$\pm$ 0.0005 & 0,8089$\pm$ 0.0005 & {\bf 0,8165$\pm$ 0.0005} \\
	\hline
\end{tabular}
\vspace{.1in}
\caption{Overall accuracy and Kappa values with standard errors for the X-ray image of a Wistar rat's jaw. The segmentation selected by Kappa as the best is shown in boldface.   }
\label{taluz}
\end{table}

\section{Summary}
\label{summary}
This paper deals with Markov random field- model based image segmentation. We discussed two different Markovian prior models and three of their most noticeable estimation algorithms, Path Constrained Viterbi Training, Iterated Conditional Modes and Graph Cut. Their presentation in a unified framework has the advantage to give better insight in their respective theoretical properties.  Graph cut was originally designed for two class images, extended later on to multimodal images. If the images do not have distinctive modes, the GC algorithm has a tendency to converge to a reduced set of classes. ICM is more robust in that aspect, since it never fails to deliver a segmentation, and it is usually better than ML. Smoothness parameter estimation is the key point in ICM implementations. Underestimation produces a dirty segmentation, so if images are known to contain a number of homogeneous patches, initialization of $\beta$ should be high, re-estimating it at each iteration until convergence.
PCVT was the only algorithm that gave a segmentation which can not be considered a smoothed version of ML. It has the capability of moving in the space of all possible segmentations away from the saddle point where ML lays.

To the knowledge of the authors, no comparison has ever been made between Hidden Potts models and Hidden Markov Mesh Models within the same computational study.

The complexity of the algorithms is quite different. Pott's ICM and Graph Cut have only one parameter to estimate and PCVT has all transitions probabilities to estimate besides the Gaussian parameters. Nevertheless, execution time has the same order in all algorithms, if PCVT is constrained to the most probable 20 sequences in all diagonals. Allowing more sequences gives more degrees of freedom for choosing the best labeling; however, the values of the probabilities of the sequences in our studies are very low, so working with more than 60 sequences increases the complexity without allowing very realistic combinations in the bag of possible sequences.
Comparing contextual segmentations of synthetic and real images with the outputs of  EM-ML and ML algorithms for independent mixture models, the three algorithms show significant improvement in terms of segmentation smoothness. It confirms the gain in dealing with spatial dependencies.

Graph Cut segmentation is the only algorithm that was not completely written by the authors.  We provide a  Matlab comprehensive toolbox \cite{Flesia2013} for PCVT and ICM algorithms, and a Graph Cut wrapper calling Boykov and Kolmogorov (2004) \cite{Boykov2004} C code,  tested with several real and synthetic images . To the author's knowledge,  our PCVT implementation is also the only distributable code that is available for 2D HMM. We will continue enhancing our toolbox since we support the idea of a general benchmark framework for Markovian segmentation methods.

\section{References}
\bibliographystyle{elsart-num}
\bibliography{biblio2}

\begin{landscape}
\begin{figure}[h!]
\centering
	\[\begin{array}{ccccc}
 & \mbox{Original} & & \mbox{Histogram}  & \\
&\includegraphics[width=5cm]{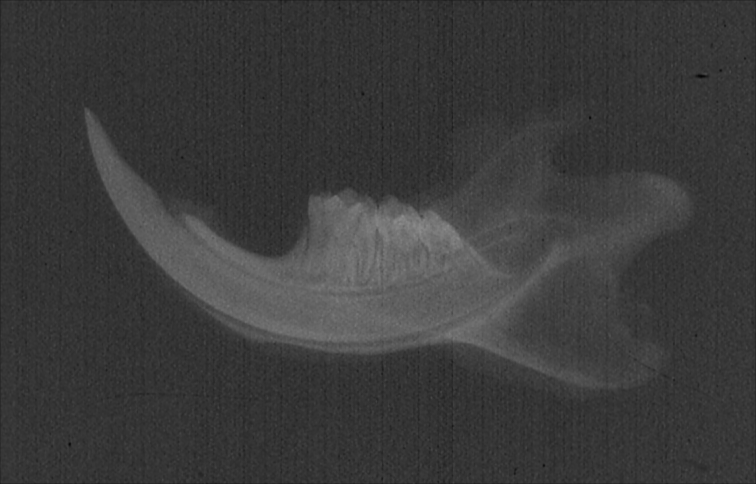}  & & \includegraphics[width=5cm]{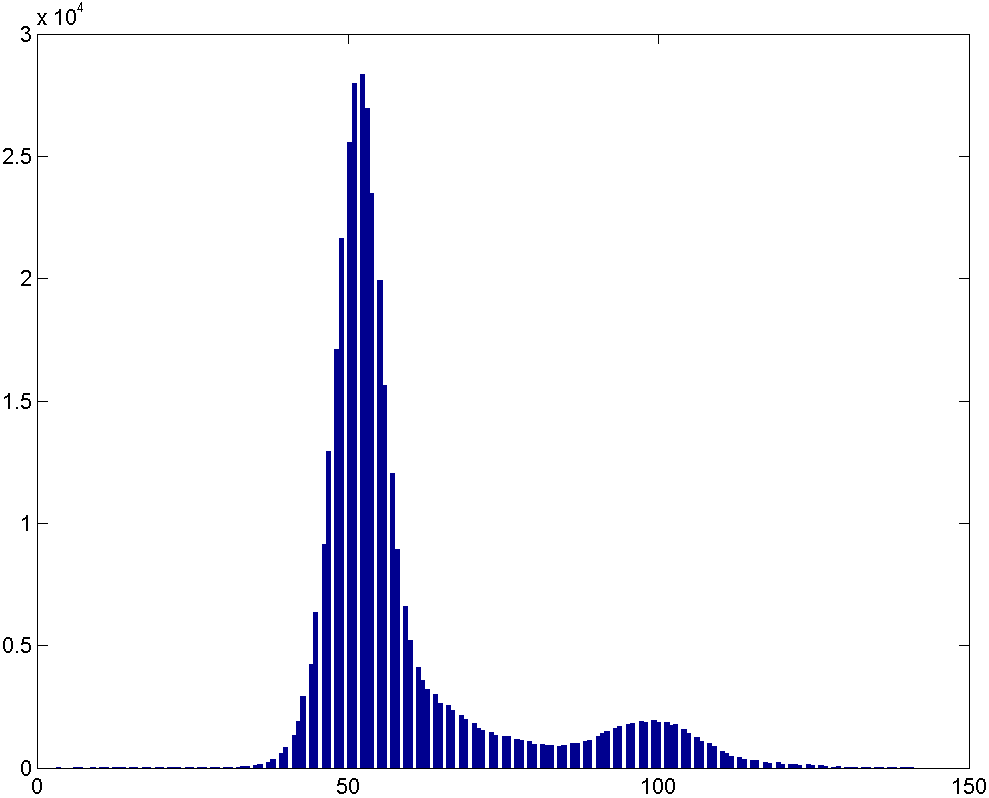}   & \\

\mbox{Exp\ a)} & \mbox{ML} & \mbox{ML-ICM} &{\bf ML-GC} & \mbox{ML-PCVT} \\
\includegraphics[width=3.2cm]{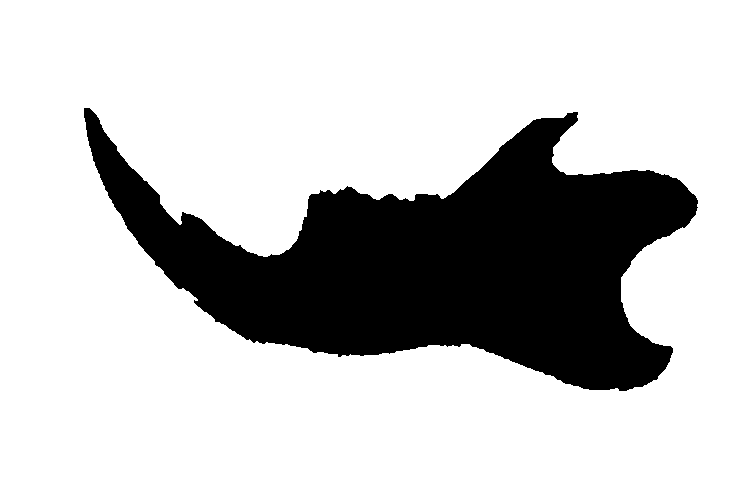} & \includegraphics[width=3.2cm]{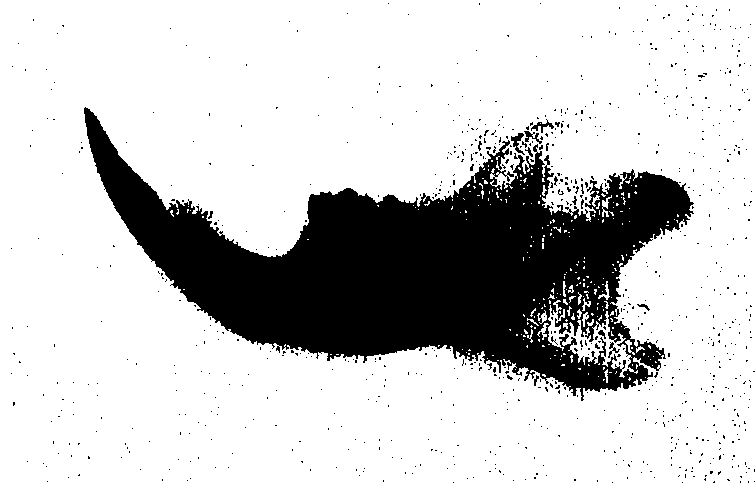} & \includegraphics[width=3.2cm]{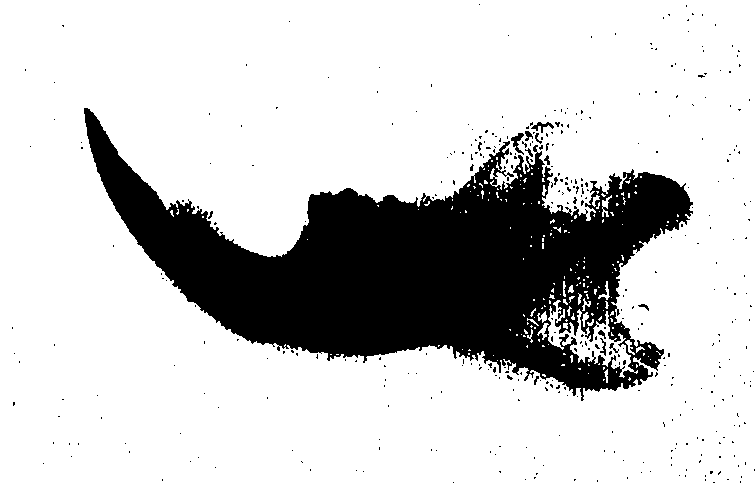} &\includegraphics[width=3.2cm]{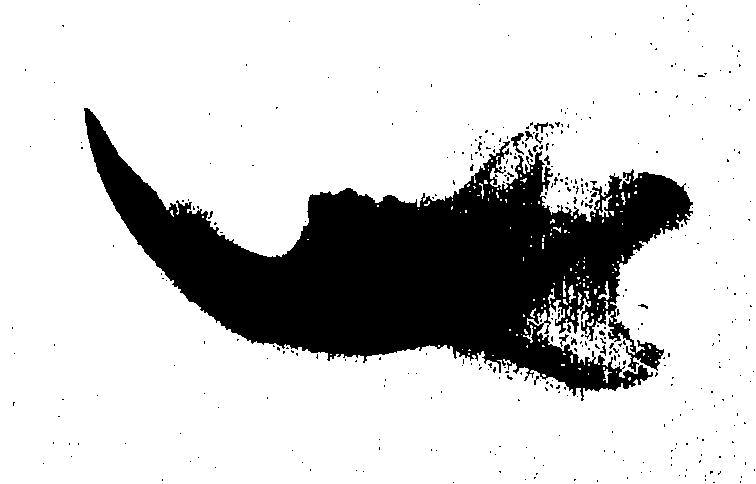} &\includegraphics[width=3.2cm]{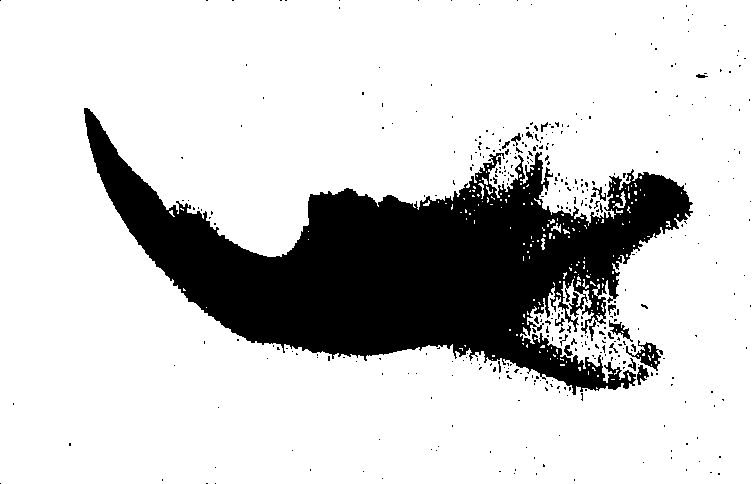} \\

\mbox{Exp\ b)} & \mbox{ML} & {\bf ML-ICM }&\mbox{ML-GC} & \mbox{ML-PCVT} \\
\includegraphics[width=3.2cm]{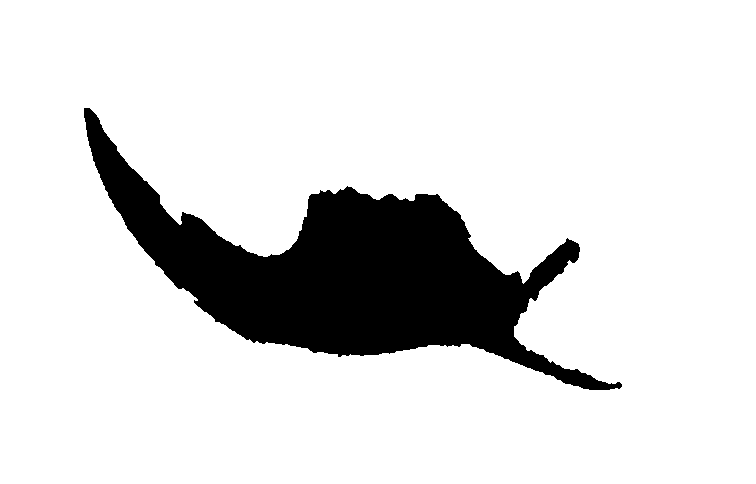} & \includegraphics[width=3.2cm]{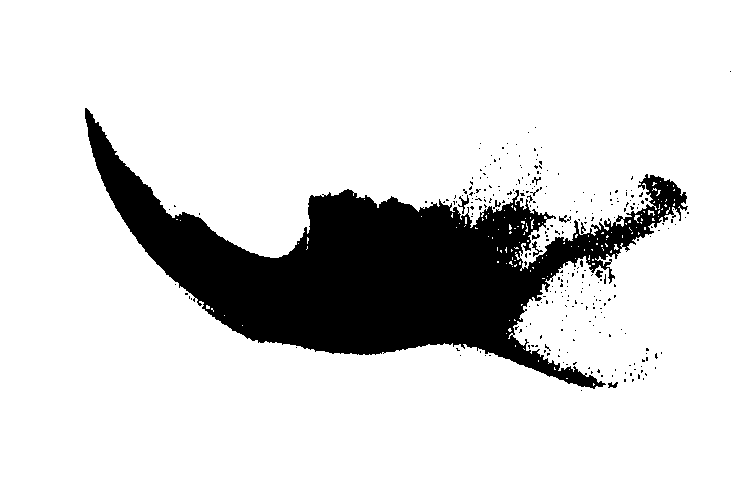} & \includegraphics[width=3.2cm]{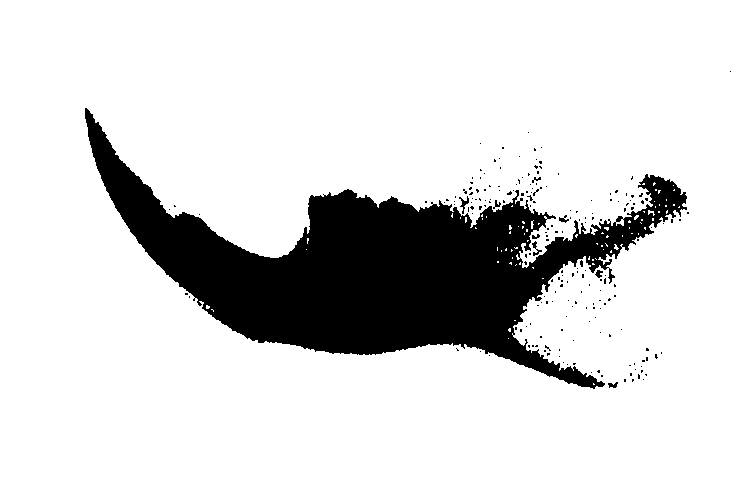} &\includegraphics[width=3.2cm]{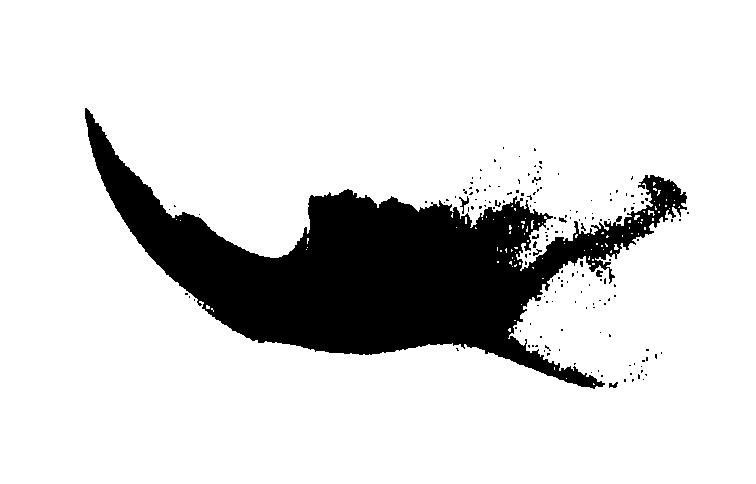} &\includegraphics[width=3.2cm]{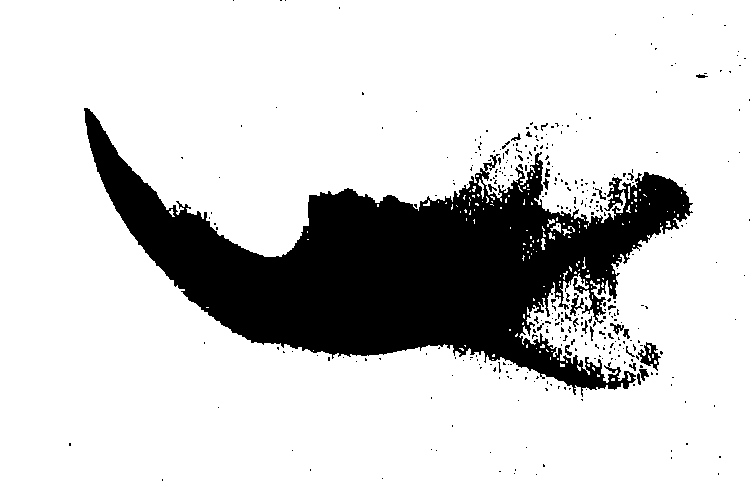} \\

\mbox{Exp\ c)} & \mbox{ML} & \mbox{ML-ICM} &{\bf ML-GC }& \mbox{ML-PCVT} \\
\includegraphics[width=3.2cm]{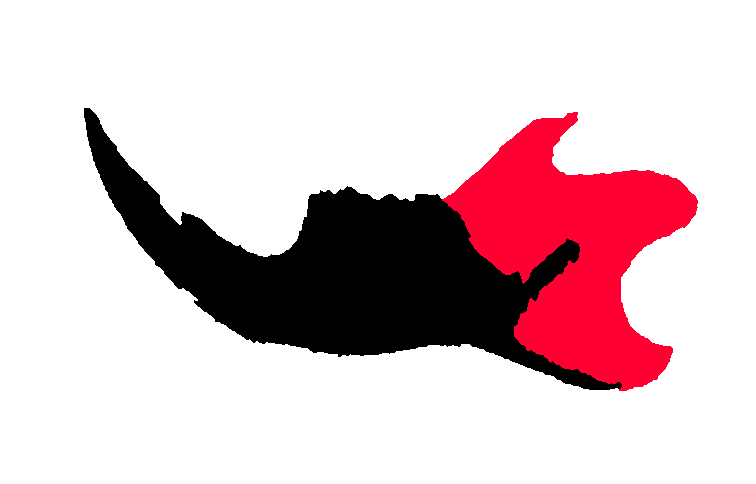} & \includegraphics[width=3.2cm]{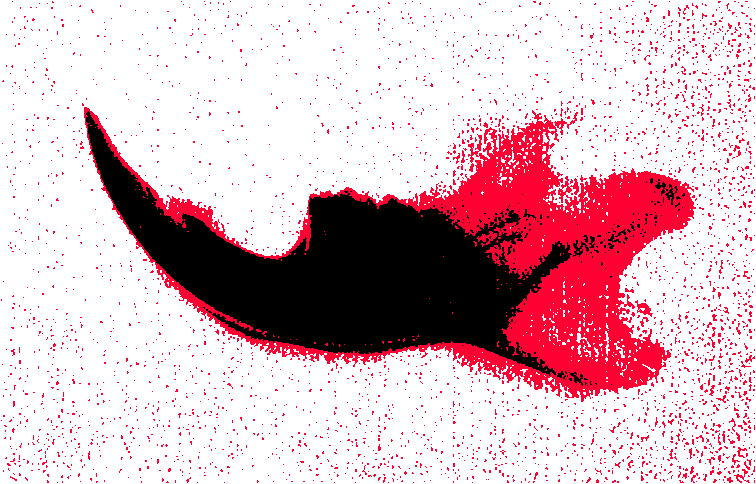} & \includegraphics[width=3.2cm]{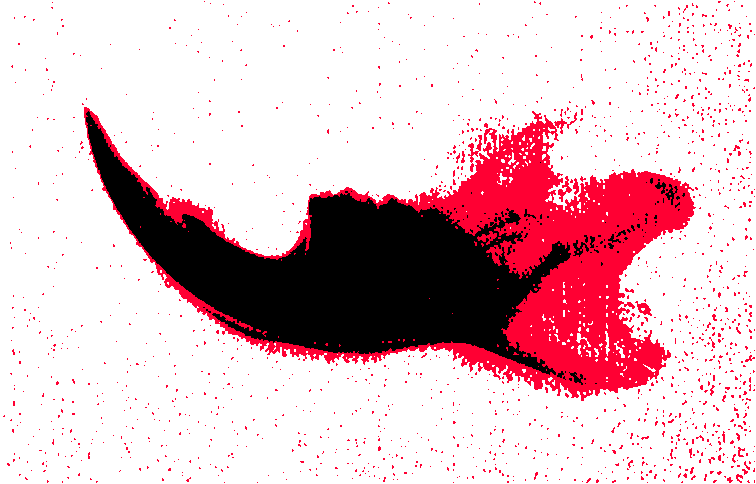} &\includegraphics[width=3.2cm]{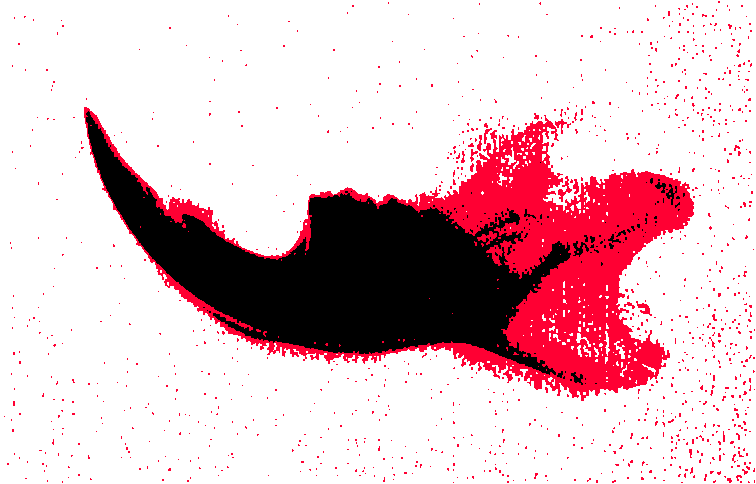} &\includegraphics[width=3.2cm]{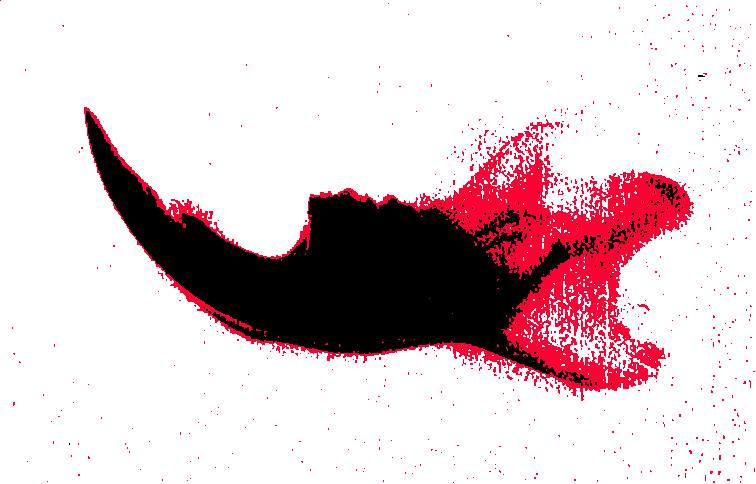} \\

\mbox{Exp\ d)} & \mbox{ML} & \mbox{ML-ICM} &\mbox{ML-GC} & {\bf ML-PCVT} \\
\includegraphics[width=3.2cm]{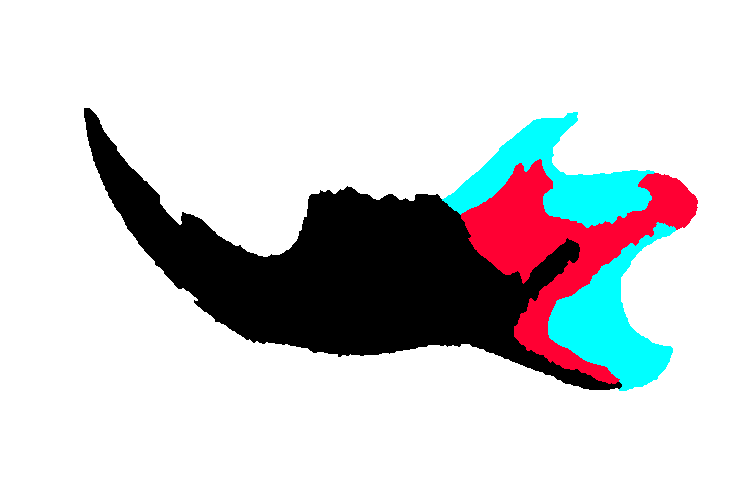} & \includegraphics[width=3.2cm]{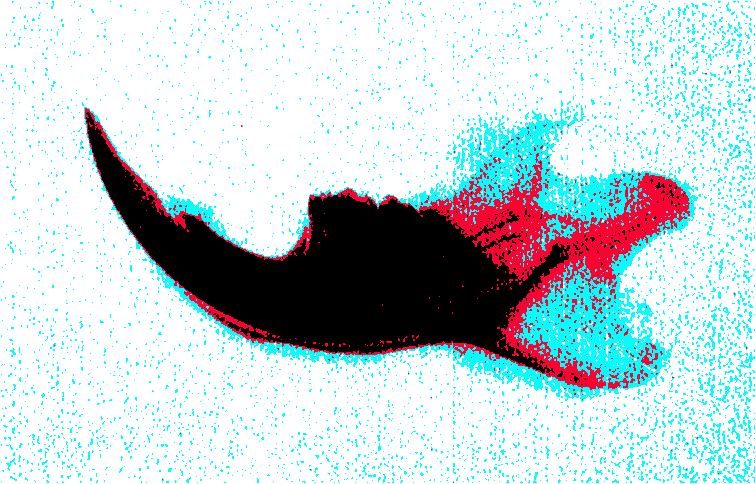} & \includegraphics[width=3.2cm]{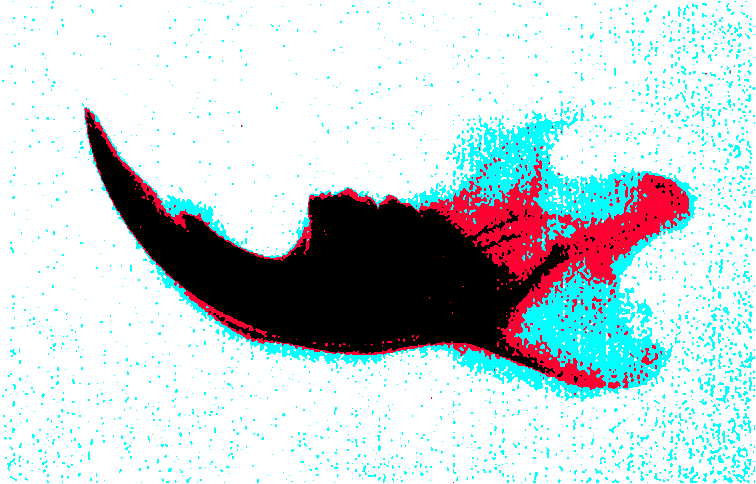} &\includegraphics[width=3.2cm]{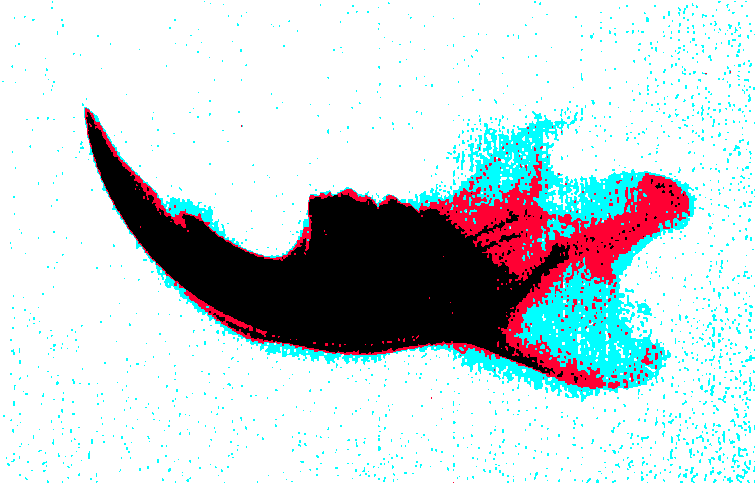} &\includegraphics[width=3.2cm]{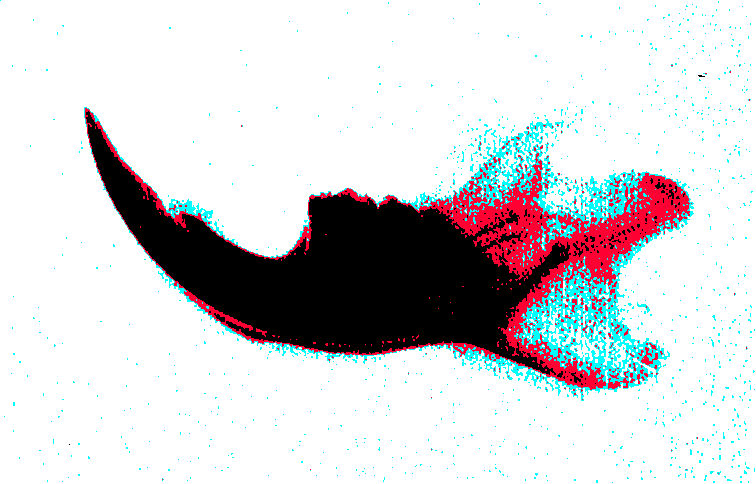} \\
	\end{array}	\]
	\caption{Segmentation results of ML, ICM, GC and PCVT for experiments (a)-(d). The segmentation selected by kappa as the best is shown in boldface. Segmentations are supervised, the initial means of the classes are selected from a small sample in the image. }
\label{seg_Luz}
\end{figure}

\end{landscape}

\begin{sidewaystable}[h!]
\centering
{\small
\begin{tabular}{|c|c|c|c|c|c|c|c|c|}
  \hline
\multicolumn{9}{|c|}{Bimodal mixture problem}\\\hline
    \multicolumn{9}{|c|}{Unsupervised problem }\\\hline
     &\multicolumn{4}{|c|}{unfiltered }&\multicolumn{4}{|c|}{filtered}\\\hline
  &\multicolumn{4}{|c|}{$(\mu,\sigma)=(60,5)$ $(\mu,\sigma)=(100,25)$ }&\multicolumn{4}{|c|}{$(\mu,\sigma)=(60,5)$ $(\mu,\sigma)=(100,25)$ }\\\hline
  &ML	&PCVT & ICM 	&{\bf GC }	                       &ML	&PCVT & ICM 	& GC\\  \hline
  OA       & 0.9232    &0.9262    &0.9840    &0.9902  & 0.9819& 0.9819    &0.9819  & 0.9819  \\
  kappa     & 0.8589$\pm$0.001    &0.8643$\pm$0.001    &0.9662$\pm$0.0007    &0.9789$\pm$0.0006 & 0.9613 $\pm$0.0007&0.9627$\pm$0.0007&  0.9614 $\pm$0.0007&  0.9611 $\pm$0.0006 \\  \hline
  &\multicolumn{4}{|c|}{$(\mu,\sigma)=(60,10)$ $(\mu,\sigma)=(100,20)$}&\multicolumn{4}{|c|}{$(\mu,\sigma)=(60,10)$ $(\mu,\sigma)=(100,20)$}\\\hline
 	 &ML&PCVT  & ICM	&GC                         &ML&PCVT  & ICM	&GC\\ \hline
OA  &0.9041    &0.9071    &0.9758    &0.9800 &0.9916 & 0.9916    &0.9916    & 0.9916 \\
 kappa&0.8289$\pm$0.001   & 0.8346$\pm$0.001  & 0.9501$\pm$0.0008    &0.9581$\pm$0.0655& 0.9821$\pm$0.0005&0.9825$\pm$0.0005&0.9820$\pm$0.0005&   0.9817$\pm$0.0003\\
  \hline
    &\multicolumn{4}{|c|}{$(\mu,\sigma)=(60,15)$ $(\mu,\sigma)=(100,15)$}&\multicolumn{4}{|c|}{$(\mu,\sigma)=(60,15)$ $(\mu,\sigma)=(100,15)$} \\\hline
&ML	&PCVT & ICM 	&GC 	  &ML	&PCVT  & ICM	&GC \\ \hline
OA        & 0.9071 & 0.9116   & 0.9882  & 0.9871&0.9958    &0.9958 &0.9958  & 0.9958    \\
kappa     & 0.8305$\pm$0.001   & 0.8290$\pm$0.0011   & 0.9747$\pm$0.0006   & 0.9722$\pm$0.0717&0.9905$\pm$0.0004& 0.9899$\pm$0.0004&0.9906$\pm$0.0004&0.9908$\pm$0.0005\\ \hline
&\multicolumn{4}{|c|}{$(\mu,\sigma)=(60,20)$ $(\mu,\sigma)=(100,10)$} &\multicolumn{4}{|c|}{$(\mu,\sigma)=(60,20)$ $(\mu,\sigma)=(100,10)$}     \\\hline
&ML	&PCVT & ICM 	&GC 	  &ML	&PCVT  & ICM	&GC \\ \hline
OA        & 0.9262 &   0.9256   & 0.9905    &0.9807&0.9883&0.9883&0.9883&0.9883\\
kappa& 0.8579$\pm$0.001   & 0.8516$\pm$0.0011   & 0.9795$\pm$0.0006    &0.9587$\pm$0.0513&0.9745$\pm$0.0006&0.9743$\pm$0.0006&0.9745    $\pm$0.0006&0.9750$\pm$0.0009\\
 \hline
  &\multicolumn{4}{|c|}{$(\mu,\sigma)=(60,25)$ $(\mu,\sigma)=(100,5)$} &\multicolumn{4}{|c|}{$(\mu,\sigma)=(60,25)$ $(\mu,\sigma)=(100,5)$}  \\ \hline
&ML	&PCVT & ICM 	&GC 	  &ML	&PCVT  & ICM	&GC \\ \hline
 OA   & 0.9510   & 0.9541    &0.9938    &0.9818& 0.9809&0.9809&0.9809&0.9809\\
 kappa & 0.9006$\pm$0.001   & 0.9052$\pm$0.001    &0.9865$\pm$0.0005   & 0.9609$\pm$0.031&0.9599$\pm$ 0.0007&0.9596$\pm$ 0.0007&0.9598$\pm$ 0.0007&0.96$\pm$ 0.0008\\\hline

\end{tabular}
\begin{tabular}{|c|c|c|c|c|c|c|c|c|}
  \hline
  \multicolumn{9}{|c|}{Supervised Problem }\\\hline
       &\multicolumn{4}{|c|}{unfiltered }&\multicolumn{4}{|c|}{filtered}\\\hline
  &\multicolumn{4}{|c|}{$(\mu,\sigma)=(60,5)$ $(\mu,\sigma)=(100,25)$ }&\multicolumn{4}{|c|}{$(\mu,\sigma)=(60,5)$ $(\mu,\sigma)=(100,25)$ }\\\hline
 &ML	&PCVT & ICM 	&GC 	  &ML	&PCVT  & ICM	&GC \\ \hline
OA       &  0.9089  &  0.9262  &  0.9129 &   0.6522&0.9917    & 0.9917&0.9917&0.9917\\
kappa    & 0.8382$\pm$0.001  &   0.8643$\pm$0.001    & 0.8442$\pm$0.001    & 0.3947$\pm$0.0022 &0.9825$\pm$0.0005&0.9641$\pm$0.0007&    0.9824$\pm$0.0005&0.9820$\pm$0.0003\\ \hline
&\multicolumn{4}{|c|}{$(\mu,\sigma)=(60,10)$ $(\mu,\sigma)=(100,20)$}&\multicolumn{4}{|c|}{$(\mu,\sigma)=(60,10)$ $(\mu,\sigma)=(100,20)$}\\\hline
OA     & 0.9057 &   0.9072   & 0.9125&    0.9125 & 0.9956& 0.9956& 0.9956& 0.9956\\
kappa   &0.8310$\pm$0.001 & 0.8346$\pm$0.001&0.8415$\pm$0.001  & 0.8414$\pm$0.001 &0.9909$\pm$0.0004&0.9827$\pm$0.0005&0.9909$\pm$    0.0004&0.9904$\pm$0.0001\\ \hline
    &\multicolumn{4}{|c|}{$(\mu,\sigma)=(60,15)$ $(\mu,\sigma)=(100,15)$}&\multicolumn{4}{|c|}{$(\mu,\sigma)=(60,15)$ $(\mu,\sigma)=(100,15)$}\\ \hline
&ML	&PCVT & ICM 	&GC 	  &ML	&PCVT  & ICM	&GC \\ \hline
OA        &  0.9087  &  0.9110  &  0.9177 &   0.9178 &0.9980&0.9980& 0.9980& 0.9980\\
kappa     & 0.8325$\pm$0.001  &   0.8278$\pm$0.0011   & 0.8467$\pm$0.001   &  0.8466$\pm$0.001 &0.9958$\pm$0.0003&0.9914$\pm$0.0004&    0.9960$\pm$0.0003&0.9956$\pm$0.0002\\ \hline
&\multicolumn{4}{|c|}{$(\mu,\sigma)=(60,20)$ $(\mu,\sigma)=(100,10)$}&\multicolumn{4}{|c|}{$(\mu,\sigma)=(60,20)$ $(\mu,\sigma)=(100,10)$}  \\\hline
OA   & 0.9274   & 0.9258  &  0.9355&    0.9358 &0.9964&0.9964&0.9964&0.9964\\
kappa & 0.8596$\pm$0.001   &  0.8521$\pm$0.001   &0.8735$\pm$0.001 &    0.8738$\pm$0.001 &0.9915$\pm$0.0004&0.9731$\pm$0.0006&0.9916$\pm$ 0.0004&0.9921$\pm$0.0006\\ \hline
  &\multicolumn{4}{|c|}{$(\mu,\sigma)=(60,25)$ $(\mu,\sigma)=(100,5)$}  &\multicolumn{4}{|c|}{$(\mu,\sigma)=(60,25)$ $(\mu,\sigma)=(100,5)$}  \\ \hline
&ML	&PCVT & ICM 	&GC 	  &ML	&PCVT  & ICM	&GC \\ \hline
 OA   &  0.9479  &  0.9542 &   0.9518&    0.9510&0.9925&0.9925&0.9925&0.9925\\
 kappa & 0.8936$\pm$0.001   &  0.9054$\pm$0.001  &   0.9008$\pm$0.001  &   0.8993$\pm$0.001 &0.9831$\pm$0.0005&0.9601$\pm$0.0007&0.9833$\pm$ 0.0005&0.9839$\pm$0.0009\\\hline
\end{tabular}
\vspace{.1in}
\caption{Segmentation results of ML, ICM, GC and PCVT for the two-circles experiment on Section \ref{two-circles}. The statistics are Overall Accuracy (OA) and Kappa with standard error. The best segmentation selected by Kappa is shown in boldface. Cases under study are filtered and unfiltered images, unsupervised and supervised segmentations. }}
\label{circ1}
\end{sidewaystable}

\begin{sidewaystable}[h!]
\centering
{\small
\begin{tabular}{|c|c|c|c|c|c|c|c|c|}
  \hline
\multicolumn{9}{|c|}{Unimodal mixture problem}\\\hline
   \multicolumn{9}{|c|}{Unsupervised Problem }\\\hline
     &\multicolumn{4}{|c|}{unfiltered }&\multicolumn{4}{|c|}{filtered}\\\hline
  &\multicolumn{4}{|c|}{$(\mu,\sigma)=(0,65)$ $(\mu,\sigma)=(60,15)$ }&\multicolumn{4}{|c|}{$(\mu,\sigma)=(0,65)$ $(\mu,\sigma)=(60,15)$ }\\\hline
 &ML	&PCVT & ICM 	&GC 	  &ML	&PCVT  & ICM	&GC \\ \hline
OA       &  0.8883 &  0.6522       & 0.9862  &  0.9392 &   0.9860  &  0.9860  &  0.9861   & 0.9863\\
kappa     &  0.7925$\pm$0.0011    & 0.3947$\pm$0.0022  &  0.9702$\pm$0.0007   & 0.8766$\pm$0.001&  0.9704$\pm$0.0007  &  0.9704$\pm$0.0007  &  0.9706$\pm$0.0007  &  0.9709$\pm$0.0007\\ \hline
&\multicolumn{4}{|c|}{$(\mu,\sigma)=(0,55)$ $(\mu,\sigma)=(60,30)$}&\multicolumn{4}{|c|}{$(\mu,\sigma)=(0,55)$ $(\mu,\sigma)=(60,30)$}\\\hline
OA       &0.8074  &  0.6522   & 0.9602  &  0.8757  &0.9924  &  0.9924  &  0.9927   & 0.9930\\
 kappa & 0.6793$\pm$0.001&0.3947$\pm$0.0022  &  0.9182$\pm$0.0009&0.7647$\pm$0.001    & 0.9835$\pm$0.0005   & 0.9836$\pm$0.0005  &  0.9842$\pm$0.0005    &0.9848$\pm$0.0005\\
  \hline
    &\multicolumn{4}{|c|}{$(\mu,\sigma)=(0,45)$ $(\mu,\sigma)=(60,45)$}  &\multicolumn{4}{|c|}{$(\mu,\sigma)=(0,45)$ $(\mu,\sigma)=(60,45)$}\\\hline
&ML	&PCVT & ICM 	&GC 	  &ML	&PCVT  & ICM	&GC \\ \hline
OA        & 0.7623   &0.6522  &  0.9249   & 0.8538 & 0.9956  &  0.9956  &  0.9960  &  0.9959\\
kappa     & 0.6297 $\pm$0.0011  &0.3947$\pm$0.0022  &  0.8543$\pm$0.0011   & 0.7299$\pm$0.001& 0.9903$\pm$0.0004   & 0.9903$\pm$0.0004    & 0.9912$\pm$0.0004   &  0.9910$\pm$0.0004 \\\hline
&\multicolumn{4}{|c|}{$(\mu,\sigma)=(0,35)$ $(\mu,\sigma)=(60,60)$}  &\multicolumn{4}{|c|}{$(\mu,\sigma)=(0,35)$ $(\mu,\sigma)=(60,60)$}  \\\hline
 OA     & 0.7168  & 0.6521 &  0.8644  &  0.7883    & 0.9935  &  0.9934   & 0.9939 &   0.9931\\
kappa    & 0.6040$\pm$0.0011  &0.3947$\pm$0.0022 & 0.7768$\pm$0.001 &0.6886$\pm$0.0009   &0.9859$\pm$0.0005  &  0.9857$\pm$0.0005  &  0.9867$\pm$0.0005  &  0.9849$\pm$0.0005\\
 \hline
  &\multicolumn{4}{|c|}{$(\mu,\sigma)=(0,25)$ $(\mu,\sigma)=(60,75)$}&\multicolumn{4}{|c|}{$(\mu,\sigma)=(0,25)$ $(\mu,\sigma)=(60,75)$} \\ \hline
&ML	&PCVT & ICM 	&GC 	  &ML	&PCVT  & ICM	&GC \\ \hline
 OA   & 0.7621  & 0.6520&  0.9038  &  0.8567&  0.9883  &  0.9888  &  0.9890  &  0.9886\\
 kappa & 0.6549$\pm$0.001 & 0.3945$\pm$0.0022  & 0.8306$\pm$0.001&0.7678$\pm$0.001& 0.9748$\pm$0.0006  &  0.9757$\pm$0.0006  &  0.9762$\pm$0.0006  &  0.9754$\pm$0.0006\\\hline
\end{tabular}
\begin{tabular}{|c|c|c|c|c|c|c|c|c|}
  \hline
  \multicolumn{9}{|c|}{Supervised Problem }\\\hline
       &\multicolumn{4}{|c|}{unfiltered }&\multicolumn{4}{|c|}{filtered}\\\hline
  &\multicolumn{4}{|c|}{$(\mu,\sigma)=(0,65)$ $(\mu,\sigma)=(60,15)$ } &\multicolumn{4}{|c|}{$(\mu,\sigma)=(0,65)$ $(\mu,\sigma)=(60,15)$ }\\\hline
  &ML	&PCVT & ICM 	&GC 	  &ML	&PCVT  & ICM	&GC \\ \hline
OA       &   0.7976  &  0.6523  &  0.7973  &  0.7946&   0.9754  &  0.9860  &  0.9784  &  0.9767\\
kappa     &   0.6350$\pm$0.0011  &  0.3950$\pm$0.0022   & 0.6343$\pm$0.0011  &  0.6299$\pm$0.0011&   0.9477$\pm$0.0008 &   0.9704$\pm$0.0007  &  0.9539$\pm$0.0008&    0.9504$\pm$0.0008\\ \hline
&\multicolumn{4}{|c|}{$(\mu,\sigma)=(0,55)$ $(\mu,\sigma)=(60,30)$}&\multicolumn{4}{|c|}{$(\mu,\sigma)=(0,55)$ $(\mu,\sigma)=(60,30)$}\\\hline
OA   & 0.8065  &  0.6522 &   0.8127  &  0.8127      & 0.9934  &  0.9924   & 0.9941 &   0.9939 \\
kappa  &   0.6777$\pm$0.001 & 0.3947$\pm$0.0022   &  0.6854$\pm$0.001  &  0.6849$\pm$0.001   & 0.9856$\pm$0.0005 &   0.9837 $\pm$0.0005  & 0.9870$\pm$0.0005  &  0.9867$\pm$0.0005\\
  \hline
    &\multicolumn{4}{|c|}{$(\mu,\sigma)=(0,45)$ $(\mu,\sigma)=(60,45)$} &\multicolumn{4}{|c|}{$(\mu,\sigma)=(0,45)$ $(\mu,\sigma)=(60,45)$} \\\hline
&ML	&PCVT & ICM 	&GC 	  &ML	&PCVT  & ICM	&GC \\ \hline
OA        &   0.7479  & 0.6522&  0.7521  &  0.7515 &   0.9955   & 0.9956   & 0.9958  &  0.9957\\
kappa     &  0.6214$\pm$0.0011 &0.3947$\pm$0.0022& 0.6263$\pm$0.0011   & 0.6253$\pm$0.0011 &   0.9903$\pm$0.0004   & 0.9904$\pm$0.0004   &  0.9909$\pm$0.0004  &   0.9906$\pm$0.0004 \\\hline
&\multicolumn{4}{|c|}{$(\mu,\sigma)=(0,35)$ $(\mu,\sigma)=(60,60)$}&\multicolumn{4}{|c|}{$(\mu,\sigma)=(0,35)$ $(\mu,\sigma)=(60,60)$}\\ \hline
OA    & 0.7255  & 0.6522&  0.7288  &  0.7281    & 0.9889  &  0.9934  &  0.9912  &  0.9918\\
kappa   & 0.6084$\pm$0.0011  & 0.3947$\pm$0.0022& 0.6123$\pm$0.0011 &   0.6115$\pm$0.0011  &0.9762$\pm$0.0006  &  0.9857$\pm$0.0005  &  0.9809$\pm$0.0005  &  0.9823$\pm$0.0005\\\hline
 &\multicolumn{4}{|c|}{$(\mu,\sigma)=(0,25)$ $(\mu,\sigma)=(60,75)$} &\multicolumn{4}{|c|}{$(\mu,\sigma)=(0,25)$ $(\mu,\sigma)=(60,75)$}  \\ \hline
&ML	&PCVT & ICM 	&GC 	  &ML	&PCVT  & ICM	&GC \\ \hline
 OA   &   0.6369  &0.6520&   0.6389  &  0.6522  &    0.9572 &  0.6522 &   0.9627 &   0.9638\\
 kappa &  0.5354$\pm$0.0013   &0.3945$\pm$0.0022&   0.5380$\pm$0.0013   & 0.3947$\pm$0.0022 &  0.9157$\pm$0.0009  &  0.3947$\pm$0.0022 &  0.9255$\pm$0.0009   &  0.9275$\pm$0.0009 \\\hline
\end{tabular}
\vspace{.1in}
\caption{Segmentation results of ML, ICM, GC and PCVT for the two-circles experiment on Section \ref{two-circles}. The statistics are Overall Accuracy (OA) and Kappa with standard error. The best segmentation selected by Kappa is shown in boldface. Cases under study are filtered and unfiltered images, unsupervised and supervised segmentations.}}
\label{circ2}
\end{sidewaystable}

\end{document}